\pdfoutput=1
\documentclass[11pt]{article}

\usepackage[final]{acl}

\usepackage{times}
\usepackage{latexsym}

\usepackage[T1]{fontenc}
\usepackage[utf8]{inputenc}

\usepackage{microtype}

\usepackage{inconsolata}

\usepackage{graphicx}

\usepackage{xspace}
\usepackage{graphicx}  

\usepackage{xcolor}

\usepackage{xurl}
\usepackage{hyperref}
\usepackage{url}
\usepackage{booktabs}
\usepackage[ruled]{algorithm2e}
\usepackage{algpseudocode}
\usepackage{enumerate}
\usepackage[english]{babel}
\usepackage{blindtext}
\usepackage{subcaption}
\usepackage{bm}
\usepackage{amsmath}
\usepackage{amssymb}
\usepackage{xspace}
\usepackage{multirow}
\usepackage{threeparttable}
\usepackage{float}
\usepackage{flafter}
\usepackage{comment}

\usepackage{ulem}
\usepackage{verbatim}
\usepackage{array}
\usepackage{enumitem}
\usepackage{booktabs}
\usepackage{multirow}
\usepackage{colortbl}
\usepackage{lipsum}
\usepackage{marvosym}
\usepackage{wrapfig}
\usepackage{pifont}

\def\modelname{\textsc{VirSci\xspace}}
\def\fig{Fig.\xspace}

\title{Many Heads Are Better Than One: Improved Scientific Idea Generation \\ by A LLM-Based Multi-Agent System}

\author{{\bf Haoyang Su}\textsuperscript{1,}\footnotemark[1], {\bf Renqi Chen}\textsuperscript{1,}\footnotemark[1], {\bf Shixiang Tang}\textsuperscript{1,5,}\footnotemark[2], {\bf Zhenfei Yin}\textsuperscript{3}, {\bf Xinzhe Zheng}\textsuperscript{1},\\
{\bf Jinzhe Li}\textsuperscript{1}, {\bf Biqing Qi}\textsuperscript{1},  {\bf Qi Wu}\textsuperscript{4}, {\bf Hui Li}\textsuperscript{4}, {\bf Wanli Ouyang}\textsuperscript{1,5}, {\bf Philip Torr}\textsuperscript{3},\\
{\bf Bowen Zhou}\textsuperscript{1,6},
{\bf Nanqing Dong}\textsuperscript{1,2,}\footnotemark[2]\textsuperscript{,}\footnotemark[3]\\
$^{1}$Shanghai Artificial Intelligence Laboratory\quad
$^{2}$Shanghai Innovation Institute\\
$^{3}$Department of Engineering Science, University of Oxford \\
$^{4}$Shanghai Institute for Science of Science\\
$^{5}$Department of Information Engineering, Chinese University of Hong Kong\\
$^{6}$Department of Electronic Engineering, Tsinghua University\\
\\
}
\begin{document}
\maketitle
{
\renewcommand{\thefootnote}{\fnsymbol{footnote}}
\footnotetext[1]{Equal contribution.}
{\fnsymbol{footnote}}
\footnotetext[2]{Corresponding authors: Nanqing Dong (dongnanqing@pjlab.org.cn) and Shixiang Tang (tangshixiang@pjlab.org.cn).}
{\fnsymbol{footnote}}
\footnotetext[3]{Project lead.}
}
%\footnotetext[2]{Corresponding author.}
%\blfootnote{Haoran Li, Dadi Guo and Donghao Li contribute equally.}

%\footnotetext[2]{Corresponding author.}
%\blfootnote{Haoran Li, Dadi Guo and Donghao Li contribute equally.}
\begin{abstract}
The rapid advancement of scientific progress requires innovative tools that can accelerate knowledge discovery. Although recent AI methods, particularly large language models (LLMs), have shown promise in tasks such as hypothesis generation and experimental design, they fall short of replicating the collaborative nature of real-world scientific practices, where diverse experts work together in teams to tackle complex problems. To address the limitations, we propose an LLM-based multi-agent system, \textit{i.e.}, \underline{Vir}tual \underline{Sci}entists (\modelname), designed to mimic the teamwork inherent in scientific research. \modelname\ organizes a team of agents to collaboratively generate, evaluate, and refine research ideas. 
Through comprehensive experiments, we demonstrate that this multi-agent approach outperforms the state-of-the-art method in producing novel scientific ideas.
We further investigate the collaboration mechanisms that contribute to its tendency to produce ideas with higher novelty, offering valuable insights to guide future research and illuminating pathways toward building a robust system for autonomous scientific discovery. The code is available at \url{https://github.com/open-sciencelab/Virtual-Scientists}.
% \textcolor{red}{We further investigate the collaboration mechanisms that contribute to its tendency to produce ideas with higher novelty, offering valuable insights to guide future research and illuminating pathways toward building a robust system for autonomous scientific discovery. The code is available at \url{https://anonymous.4open.science/r/VIRSCI}.}

\end{abstract}
\addtocontents{toc}{\protect\setcounter{tocdepth}{0}} % 设置目录深度为 0，即不包含正文部分的条目
\section{Introduction}
\label{sec:intro}
The rapid acceleration of scientific research necessitates innovative tools for exploring new concepts and tackling complex challenges~\citep{park2023papers}. Automatic scientific discovery has emerged as a promising solution to accelerate innovation, aligning with the ultimate goal within the scientific community~\citep{langley1987scientific}.
With the development of artificial intelligence (AI), automatic scientific discovery has the potential to revolutionize how research is conducted by automating key steps in the scientific process, ranging from hypothesis generation to experimental design~\citep{raghu2020survey, spangler2014automated}.

Recent works like AI Scientist~\citep{lu2024ai}, ResearchTown~\cite{yu2024researchtown}, and HypoGen~\cite{qi2024large} leverage LLMs~\citep{openai,llama3} as intelligent agents to simulate the scientific idea generation process and advance automatic scientific discovery at various stages, including literature review~\citep{shi2023towards,hsu2024chime} and experimental design~\citep{wu2023autogen,huang2024crispr}. 
However, these efforts either rely on a \textbf{single-agent} system, overlooking the collaborative nature of real-world research~\citep{kayacik2019identifying, gauch2003scientific, linsey2005collaborating}, or employ an oversimplified collaboration framework and unrealistic data (\emph{e.g.} manually crafted personal profiles, synthetic collaboration networks) to model a \textbf{multi-agent} system, failing to capture the dynamic relationships that characterize real scientific teams. Consequently, they provide limited insights into multi-agent collaboration, which is essential for advancing autonomous scientific discovery.

% \textcolor{red}{Recent works like AI Scientist~\citep{lu2024ai}, ResearchTown~\cite{yu2024researchtown}, and HypoGen~\cite{qi2024large} leverage LLMs~\citep{openai,llama3} as intelligent agents to simulate the scientific idea generation process and advance automatic scientific discovery at various stages, including literature review~\citep{shi2023towards,hsu2024chime} and experimental design~\citep{wu2023autogen,huang2024crispr}. 
% However, these efforts either rely on a \textbf{single-agent} system, overlooking the collaborative nature of real-world research~\citep{kayacik2019identifying, gauch2003scientific, linsey2005collaborating}, or employ an oversimplified collaboration framework and unrealistic data (\emph{e.g.} personal information, collaboration networks) to model a \textbf{multi-agent} system, failing to capture real-world complexities. Consequently, they provide limited insights into multi-agent collaboration, which is essential for advancing autonomous scientific discovery.}

\begin{figure*}[!t]  
    \centering  
    \includegraphics[width=0.92\textwidth]{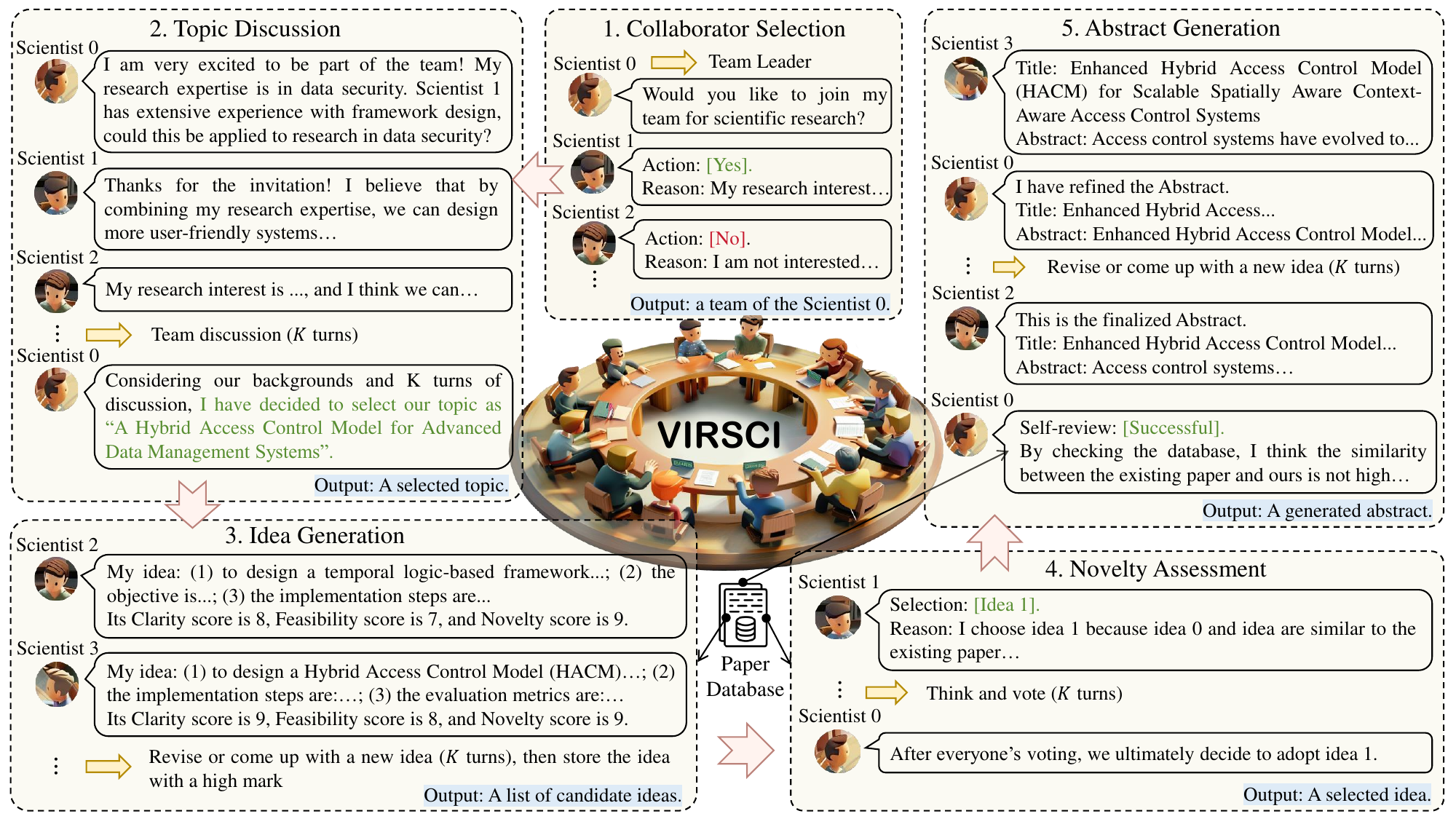} 
    \vspace{-1mm}
    \caption{The proposed LLM-based multi-agent system, \modelname, includes five key steps: \textit{Collaborator Selection}, where a research team is assembled; \textit{Topic Discussion}, where the research topic is determined; \textit{Idea Generation}, where team members propose and refine ideas; \textit{Novelty Assessment}, where ideas are evaluated and voted on to select the best one; and \textit{Abstract Generation}, where the selected idea is developed into a complete abstract.}
    \label{fig:example_image}
\end{figure*}

To address these limitations, we build a virtual \textit{Scientific Research Ecosystem} to benchmark the potential of multi-agent systems in scientific idea generation. Specifically, \textit{Scientific Research Ecosystem} is a digit twin of research communities at the given time point, \textit{e.g.,} Jan 1st, 2024. It have three components: (1) virtual scientists whose backgrounds and publications are cloned from real scientists at that moment, (2) a past paper database where papers are published before that moment and (3) a contemporary paper database where papers are published after that moment. To evaluate the novelty of the generated ideas, we introduce three metrics from different perspectives: dissimilarity to past papers, alignment with research trends, and the potential influence on contemporary research~\citep{shao2020bert, yang2022gender}.
By comparing the generated abstract against two paper databases, we ensure the generated ideas are innovative and align with emerging scientific directions, validating the effectiveness of our approach. Real examples and their professional analysis are shown in Appx.~\ref{sec:discussion_feasibility}.

Based on the virtual \textit{Scientific Research Ecosystems}, we propose an LLM-based multi-agent system, \underline{Vir}tual \underline{Sci}entists (\modelname), designed to harness the potential of LLM agents in assisting autonomous scientific idea generation. Leveraging the inherent human-like reasoning capabilities of LLMs~\citep{xie2024can}, \modelname\ simulates the collaborative process of generating scientific ideas~\citep{perry2017creativity,muzzio2024collaborative}, by the following five steps (see \fig\ref{fig:example_image}):
(1) \textit{Collaborator Selection},
(2) \textit{Topic Selection},
(3) \textit{Idea Generation},
(4) \textit{Idea Novelty Assessment}, and
(5) \textit{Abstract Generation}.
In \textit{Collaborator Selection} stage, given a randomly selected agent as the team leader, it will exploit historical co-authors based on its collaboration history and academic social networks, while also exploring potential collaborators whose expertise and research interests align with the team's goals~\cite{march1991exploration,rzhetsky2015choosing,zeng2019increasing}.
In \textit{Topic Selection} stage, the scientists will discuss topics of common interest. The discussion will be terminated and restarted if a consensus cannot be achieved among the majority of the team. The scientists who are not interested can choose to leave the discussion at will. Otherwise, the discussion continues until a final topic is determined.
In the \textit{Idea Generation} stage, the virtual scientists retrieve relevant papers from the past paper database and engage in both inter- and intra-team discussions, where collaborators participate in iterative dialogues based on their backgrounds. This inter- and intra-team discussion distinguishes our approach from previous group discussion patterns~\cite{zhang2023exploring,qi2024large,qian2024chat}, enabling agents within the team to proactively seek advice from external agents (inter-team) through an ``Invitation Mechanism'', while effectively balancing diverse perspectives within the team (intra-team). 
Once the discussions are over, the \textit{Abstract Generation} stage begins, and the team generates a comprehensive abstract that encapsulates the proposed ideas.

We conduct extensive experiments to verify the effectiveness of \modelname\ in producing novel scientific ideas on both single-discipline and multi-discipline datasets.
The findings prove that the multi-agent system improves the single-agent executive by, on average, \textbf{+13.8\%} and \textbf{+44.1\%} in alignment with and potential impacts on contemporary research, respectively.
Furthermore, our experiments investigate collaboration mechanisms among agents that influence the performance of idea generation. The patterns observed in the experimental results align with findings from prior Science of Science studies~\citep{fortunato2018science, wu2019large, zeng2021fresh, shi2023surprising} published in Top venues, \emph{e.g.,} \textit{Science} and \textit{Nature}, providing valuable insights to guide future research toward autonomous scientific discovery.
Our core contributions are as follows:

1) To the best of our knowledge, we propose the first multi-agent system with a scientific research ecosystem for conducting and benchmarking scientific collaborations, named \modelname, where real data is used for role-playing and objective evaluation.

2) To simulate a reliable scientific collaboration process, we propose an end-to-end pipeline that spans team organization to idea generation. A novel inter- and intra-team discussion mechanism is introduced to promote communication topology and enhance the simulation realism.

3) Extensive experiments demonstrate that multi-agent collaboration improves outcome quality, surpassing existing methods. We also conduct systematic experiments to investigate factors influencing the system, with findings aligning with established principles from the Science of Science, such as fresh teams tend to generate more innovative research, underscoring \modelname's reliability as a powerful tool for autonomous scientific discovery.

% 1) \textcolor{red}{To the best of our knowledge, we propose the first multi-agent system with a scientific research ecosystem for conducting and benchmarking scientific collaborations, named \modelname, where real data is used for role-playing and objective evaluation.}

% 2) \textcolor{red}{To simulate a reliable scientific collaboration process, we propose an end-to-end pipeline that spans team organization to idea generation. A novel inter- and intra-team discussion mechanism is introduced to promote communication topology and enhance the simulation realism.}

% 3) \textcolor{red}{Extensive experiments demonstrate that multi-agent collaboration improves outcome quality, surpassing existing methods. We also conduct systematic experiments to investigate factors influencing the system, with findings aligning with established principles from the Science of Science, such as fresh teams tend to generate more innovative research, underscoring \modelname's reliability as a powerful tool for autonomous scientific discovery.}
\section{Related Work}
\label{sec:related-work}

\subsection{AI for Scientific Discovery}
In recent years, AI has transformed scientific discovery by offering powerful tools to enhance research processes~\citep{xu2021artificial}. 
Generative AI, in particular, accelerates basic scientific discoveries by tackling complex tasks such as molecular identification~\citep{vignac2022digress}, protein structure prediction~\citep{abramson2024accurate}, and proteomics research~\citep{ding2024automating}, significantly reducing experimental iteration times.
% In recent years, AI has fundamentally reshaped the landscape of scientific discovery by providing powerful tools that enhance various research processes~\citep{xu2021artificial}.
% Especially, generative AI techniques can accelerate basic scientific discoveries by tackling complex tasks such as molecular identification~\citep{vignac2022digress}, protein structure prediction~\citep{abramson2024accurate}, and proteomics research~\citep{ding2024automating}, significantly reducing the time needed for experimental iterations.
% These advancements have found wide application across diverse fields such as chemistry~\citep{liu2023multi,ding2024automating}, meteorology~\citep{bi2023accurate}, and medicine~\citep{rajpurkar2022ai}, \etc.
Besides, with the advent of LLMs, AI methodologies can step further and collaborate in streamlining critical stages of the scientific pipeline, including hypothesis generation, experimental design, data acquisition, and analysis\citep{zheng2023large, wang2023scientific, miret2024llms, wysocki2024llm, lu2024ai, si2024can}.
Nevertheless, these approaches lack the collaborative nature of the scientists intrinsic to real-world research.
\modelname\ is the first to harness the power of an LLM-based multi-agent system to facilitate the generation of research ideas in autonomous scientific discovery.

\subsection{Collaboration in Multi-Agent Systems}
A multi-agent system for team collaboration leverages autonomous agents to coordinate, communicate, and solve tasks within a shared environment, simulating human teamwork dynamics~\citep{dorri2018multi}. Traditional systems typically rely on semi-autonomous agents using explicit protocols and structured messages to achieve shared goals~\citep{dunin2011teamwork, bakliwal2018multi}.
The emergence of LLMs has reshaped this landscape by enabling agents to communicate in natural language, fostering more intuitive and flexible interactions~\citep{park2023generative}. Studies have demonstrated that LLM-based multi-agent systems outperform single-agent setups in tasks such as programming, game playing, and complex reasoning~\citep{du2024multi, wang2024rethinking, light2023avalonbench}. 
However, prior multi-agent frameworks~\citep{qi2024large,yu2024researchtown} for focused scientific idea generation (1) lack dynamic team composition, restricting team members from accessing insights beyond their initial group, and (2) do not incorporate real scientist data and collaboration relationship, leading to impractical conclusions. To overcome these limitations, we introduce an inter- and intra-team discussion mechanism, enabling flexible collaboration among virtual scientists (supported by real data) and fostering the generation of \textit{de novo} scientific ideas.
% \textcolor{red}{However, prior multi-agent frameworks~\citep{qi2024large,yu2024researchtown} for focused scientific idea generation (1) lack dynamic team composition, restricting team members from accessing insights beyond their initial group, and (2) do not incorporate real scientist data and collaboration relationship, leading to impractical conclusions. To overcome these limitations, we introduce an inter- and intra-team discussion mechanism, enabling flexible collaboration among virtual scientists (supported by real data) and fostering the generation of \textit{de novo} scientific ideas.}

% In this work, we strategically implement the power of LLM-based multi-agent systems and propose a novel ``Invitation Mechanism'' to function as collaborative scientists, promoting \textit{de novo} scientific ideas.

\section{The \modelname}
\label{headings}

% \subsection{Overview}
\begin{figure*}[!t]  
    \centering  
    \includegraphics[width=0.92\textwidth]{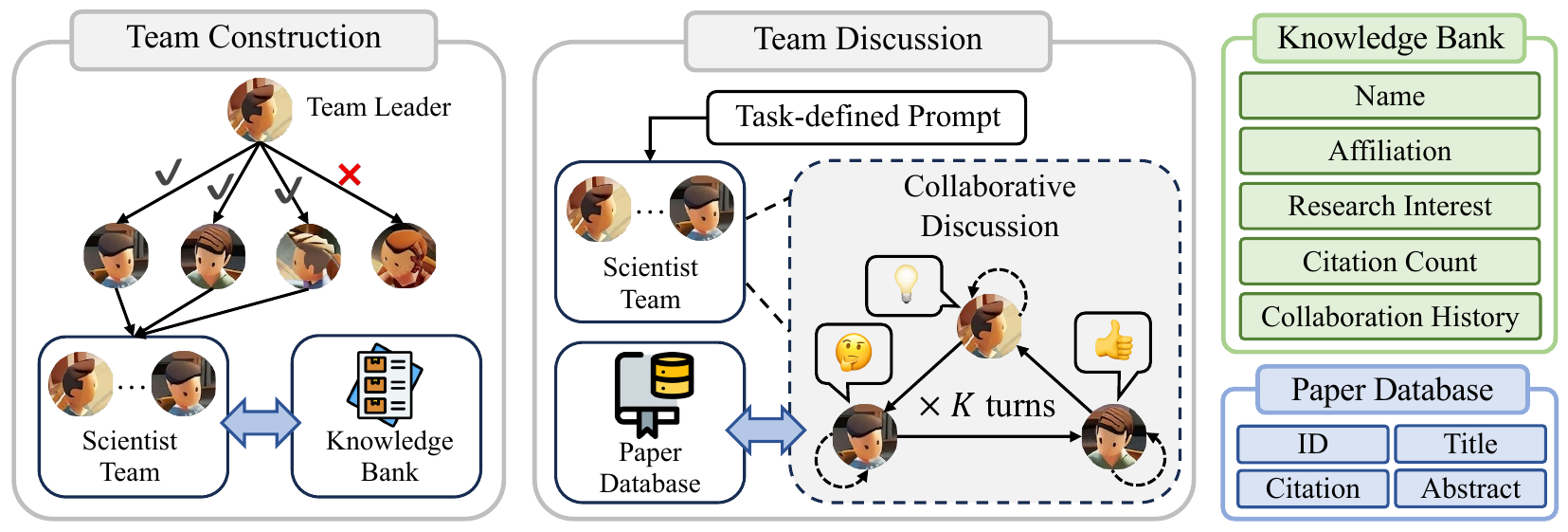}
    \vspace{-1mm}
    \caption{Key components of the proposed system. The left section illustrates the collaborator selection process, where the team leader forms a research team. The middle section highlights the discussion routine, a fundamental part of every step in the system, where the team engages in collaborative dialogue to progress through tasks. The right section depicts the architecture of the author knowledge bank and paper database, which provide critical information used throughout the collaboration process.}
    \label{fig:framework_image}
\end{figure*}

In this paper, we aim to build a multi-agent system using real-world academic datasets to simulate how a scientist assembles a research team and collaboratively generates an abstract that details a novel scientific idea\footnote{An abstract effectively represents the key aspects of scientific research and serves as a concise
reflection of its novelty. Additionally, considering the computational constraints, we focused our evaluation primarily on the generated abstracts rather than the full texts. This allows us to assess the system's contributions while efficiently
utilizing available resources.
}. Our \modelname\ system consists of two components: a scientific research ecosystem and a multi-agent system for scientific collaboration and idea generation. 

\subsection{The Scientific Research Ecosystem}
\label{Database_preparation}
\label{scientist_information}
The scientific research ecosystem comprises two main components: paper information and corresponding author information ranging from year $y_{\text{start}}$ to $y_{\text{end}}$. First, we select a year $y_{\text{bound}}$ as a time point and split the papers into two subsets: past papers $B_{\text{past}}$ and contemporary papers $B_{\text{con}}$. We further extract authors from $B_{\text{past}}$ to form the complete set of scientists $S$, with each scientist's background information stored in the author knowledge bank, and the adjacency matrix $A$, which represents the collaboration counts between scientists. 

\noindent \textbf{Past Paper Database.}~To construct the past paper database $B_{\text{past}}$ using the Faiss\footnote{Faiss is a Python library designed for efficient similarity search and clustering of dense vectors.}~\cite{johnson2019billion}, we selected papers published before the $y_{bound}$. Each paper includes essential information such as its title, citation count, and abstract.

\noindent \textbf{Contemporary Paper Database.}~The contemporary paper database $B_{\text{con}}$, also constructed with Faiss, consists of papers published after $y_{bound}$. Similarly, each paper's basic information is structured in the same way as the past papers. Although using papers from this time range may raise concerns about data leakage, given that LLMs are trained on data within this period, we will explain in detail why this does not pose a threat to the overall validity of our experiments in Appx.~\ref{appx:data_leakage}.

\noindent \textbf{Author Knowledge Bank.}~For each scientist in $S$, we extract their basic profile from the pre-processed dataset (More details are shown in Sec.~\ref{sec:experimental_settings}), which includes their name, affiliations, citation count, research interests, and collaboration history. Using the KnowledgeBank module from AgentScope~\citep{agentscope}, we embed these scientist profiles into the author knowledge bank. This allows agents to quickly access and familiarize themselves with other initialized agents’ information. Notably, real author names are masked to prevent data leakage and privacy problems during agent initialization (See Sec.~\ref{ethics}).

\noindent \textbf{Adjacency Matrix.}~Given the scientist set $S$, let $A$ represent the adjacency matrix, where $A_{i,j}$ denotes the number of times that scientist $i$ has collaborated with scientist $j$. To prevent agents from always choosing previously collaborated scientists, overlooking the benefit of fresh collaborations that often lead to more original and impactful research~\citep{zeng2021fresh}, we increment all values in $A$ by 1 (more experiments about the increment function in Appx.~\ref{app_initialization}). 
This adjustment ensures an explore-exploit mechanism, allowing scientists with no prior collaborations a chance to be selected, thereby encouraging agents to explore new partnerships.

\subsection{The Multi-agent System}
We first randomly sample a scientist $s_{0}$ from $S$ as the team leader. The team leader then follows these steps to produce an abstract: (1) \textit{Collaborator Selection}, (2) \textit{Topic Discussion}, (3) \textit{Idea Generation}, (4) \textit{Idea Novelty Assessment}, and (5) \textit{Abstract Generation}. To help each agent become familiar with the backgrounds of other team members without overloading the initialization prompt, we employ retrieval-augmented generation (RAG)~\citep{lewis2020retrieval}, used throughout all five steps. All necessary prompts and example scenarios are shown in Appx. \ref{sec:prompts} and \ref{sec:examples}.

\noindent \textbf{Collaborator Selection.}~The first step in our system is to form a team of scientists, $T = \{s_{0}, \dots, s_{i}, \dots, s_{n}\}$, where $n$ denotes the team size. When $s_{0}$ is selecting collaborators, we convert the adjacency matrix, $A$, into a probability distribution using the following equation: $P_{i,j} = \frac{A_{i,j}}{\sum_{j=1}^{N} A_{i, j}}$, where $N$ denotes the size of $S$. This allows the team leader to iteratively send invitations to preferred collaborators. Upon receiving an invitation, the invited scientist evaluates whether to join the team using the chain-of-thought process~\citep{wei2022chain}, considering the profiles of $s_0$ and the current team members. If accepted, the scientist is added to the team $T$. This process continues until the pre-defined team size $n$ is reached.

\noindent \textbf{Topic Discussion.}~The next step is to propose a research topic, which will guide the research direction. Inspired by multi-round collaboration~\citep{mezirow2003critical,sunstein2005societies,amgoud2009using} and multi-agent collaboration strategies~\citep{xu2023magic,zhang2023exploring,shinn2024reflexion}, we design an inter- and intra-team discussion mechanism. In this mechanism, team members engage in discussions based on a specific task description prompt (inter-team discussion). The default discussion mechanism between agents follows a round-table format with a sequential progression (Additional discussion topologies are presented in Appx.~\ref{sec:ablation_topologies}). This process is also applied to subsequent collaboration steps. While allowing agents to decide when to stop the discussion would better reflect real-world scenarios, fixing the number of turns ensures consistent inference costs across different team settings in our experiments. Therefore, we leave the discussion of adaptive turn numbers to the ablation study (See Appx.~\ref{sec:ablation_adaptive}).
The prompt for agent $i$ during the topic discussion is:
\begin{equation} 
    Q_{k,i}=\langle Q_{team},Q_{topic},\overset{k-1}{\underset{t=1}{\bigcup}}(\overline{D_{t}}),\overset{i-1}{\underset{j=0}{\bigcup}}(R_{k,j})\rangle,
\end{equation}
where $Q_{team}$ denotes the description of the current team members, $Q_{topic}$ represents the task description for the topic discussion, $R_{k,j}$ is the response of agent $j$ at turn $k$, and $\overline{(D_{t})}$ is the team leader's summary of dialogues from turn $t$, where $D_{t}=\{R_{t,0},R_{t,1},\dots,R_{t,n}\}$. 
Given the prompt $Q_{k,i}$, each scientist agent generates a response $R_{k,i}$, sampled from a probability distribution $R_{k,i} \sim P_{s_{i}}(\cdot|Q_{k,i})$.
Since agents can use RAG to access the author knowledge bank during discussions, they may seek advice from scientists who are relevant to the topic but not part of the team. In such cases, we initialize a new agent with the mentioned scientist's profile and include their responses in the discussion (intra-team discussion). However, to maintain the fixed team size, this agent is not added to the team. This process is termed the \textbf{``Invitation Mechanism''} and is also applied in subsequent steps, with its effectiveness demonstrated in Appx.~\ref{sec:component}. An example scenario is shown in Appx.~\ref{sec:invitation}. 
After $K$ turns of discussion, the team leader generates the final research topic $R_{topic}$ when the team reaches a consensus (otherwise the discussion is terminated and restarted), based on the content: $\langle Q_{topic}, \overset{K-1}{\underset{t=1}{\bigcup}}(\overline{D_{t}}), \overset{n}{\underset{j=0}{\bigcup}}(R_{K,j})\rangle$. At last, each scientist will be asked if they are interested in the topic. If not, they may leave the team; otherwise, they can stay and continue future discussions.

\noindent \textbf{Idea Generation.}~Third, the team is tasked with proposing several potential ideas. To align with genuine research workflows and mitigate LLM illusions~\citep{huang2023survey}, each agent is required to generate a comprehensive response that includes three key components: (1) a description of the idea, (2) a specific experimental plan, and (3) a self-assessment covering metrics such as novelty, feasibility, and clarity, representing the agent’s confidence (See Appx. Fig.~\ref{fig:idea_prompt}).

At the start of the idea generation process, when no ideas have yet been proposed, the agent is provided with references by searching $B_{past}$ using the topic $R_{topic}$, denoted as $B_{past}(R_{topic})$. The first prompt is defined as: $Q_{1,0}=\langle Q_{idea},R_{topic},B_{past}(R_{topic})\rangle$,
where $Q_{idea}$ represents the task description. 

Inspired by the concept of gradually expanding an archive of ideas~\citep{zhang2023exploring,lu2024ai}, when a scientist $s_{i}$ at turn $k$ receives an existing idea from the previous response $R_{k,i-1}$, we retain the previously generated ideas along with their corresponding references from $B_{past}$. These are passed to the next agent, who can either refine the existing idea or propose a new one, depending on its choice. The prompt $Q_{k,i}$ is represented as:
\begin{equation}
    \langle Q_{idea},R_{topic}, B_{past}(R_{k,i-1}), \overset{k-1}{\underset{t=1}{\bigcup}}(\overline{D_{t}}), \overset{i-1}{\underset{j=0}{\bigcup}}(R_{k,j})\rangle.
\end{equation}

Afterwards, the response of $S_i$ at turn $k$ can be represented as $R_{k,i}\sim P_{s_{i}}(\cdot|Q_{k,i})$. After $K$ turns of discussion, we retain the three ideas with the highest confidence and store them in the idea list $I$.

\noindent \textbf{Novelty Assessment.}~To enhance the quality of ideas and mitigate agent overconfidence, we introduce an idea novelty assessment, enabling agents to compare each idea with related papers from $B_{past}$ and vote for the idea they consider most novel. Given the idea list $I$, agents search for related papers using each idea's description to determine whether it significantly overlaps with existing works. To simulate a blind review process, no dialogue memory is included in the prompt. The prompt for $s_{i}$ at turn $k$ is defined as:
\begin{equation}
    Q_{k,i}=\langle Q_{check},\overset{3}{\underset{j=1}{\bigcup}}(I_{j},B_{past}(I_{j})\rangle,
\end{equation}
where $I_{j}$ is the $j$-th idea in $I$. Following chain-of-thought process, the response $R_{k,i}\sim P_{s_{i}}(\cdot|Q_{k,i})$ includes the scientist’s preferred idea and the reasoning behind their choice. The idea receiving the highest votes is then selected as the final idea, $R_{idea}$, for abstract generation. The effectiveness of novelty assessment is discussed in Appx.~\ref{sec:component}.

\noindent \textbf{Abstract Generation.}~Lastly, the team is required to produce a comprehensive abstract that includes the following sections: (1) Introduction, (2) Objective, (3) Methods, (4) Expected Results, and (5) Conclusion~\citep{alexandrov2007writing}. At the start of abstract generation, the team leader provides an initial draft based on $R_{idea}$. The first abstract-generation prompt is: $Q_{1,0}=\langle Q_{abstract},R_{idea}\rangle$, where $Q_{abstract}$ represents the task description and format requirements.

When an abstract is provided by the previous response $R_{k,i-1}$, the next scientist’s response should include: (1) an evaluation of the prior abstract (evaluation metrics are detailed in Appx. Fig.~\ref{fig:second_abstract_prompt}), (2) proposed modifications, and (3) the revised abstract to enable continuous refinement. The prompt is: \begin{equation}\label{origin_abstract}
    Q_{k,i}=\langle Q_{abstract},Q_{judgment},R_{k,i-1}\rangle,
\end{equation}
where $Q_{judgment}$ is the prompt that asks agents to evaluate the previous abstract. Dialogue history is not included in this prompt since the process is iterative and focuses on refining a single abstract. Including previous versions would make the prompt redundant. After $K$ turns of revision, the final abstract is denoted as $R_{abstract}$.

A self-review mechanism is also considered after $R_{abstract}$ is finalized to pre-check its novelty. The optimized abstract $R_{abstract}$ is provided to the team leader to assess novelty by comparing it to similar papers in $B_{past}$ (See Appx.~\ref{sec:self_review} for more details). Because it introduces uncertainty in total inference cost, making it difficult to ensure fair experimental comparisons, we only discuss the effectiveness of this module in the ablation study (See Appx.~\ref{sec:component}).

\section{Empirical Study}
\subsection{Experimental Settings}
\label{sec:experimental_settings}
\noindent \textbf{Datasets.}~We evaluate the performance of \modelname\ and baseline methods on two datasets: the AMiner Computer Science dataset\footnote{\url{https://www.aminer.cn/aminernetwork}}, containing 1,712,433 authors and 2,092,356 papers from 1948 to 2014, and the Open Academic Graph 3.1\footnote{\url{https://open.aminer.cn/open/article?id=65bf053091c938e5025a31e2}}, comprising 35,774,510 authors and 130,710,733 papers as of 2023. For quality assurance, we applied filtering strategies, reducing the Computer Science dataset to 156 authors and 178,197 papers, and the Open Academic Graph to 3,169 authors and 201,131 papers. Additional dataset details and filtering strategies are provided in Appx.~\ref{appx:experimental_settings}.

\noindent \textbf{Evaluation Metrics.}~
% Unlike previous work~\citep{qi2024large,lu2024ai} that primarily relied on LLM-based scoring as the evaluation criterion, 
% In this paper, we aim to adopt a more objective approach as the metric. Since no single evaluation metric perfectly captures the novelty of scientific outputs, we employ three common metrics that align with our intuition: 
In this paper, we adopt a more objective approach by using three common metrics that align with our intuition, as no single metric fully captures the novelty of scientific outputs:
(1) \textit{Historical Dissimilarity (HD)}: The average Euclidean distance between the generated abstract embedding and embeddings of the 5 most similar abstracts in $B_{\text{past}}$~\citep{shao2020bert, zhou2024fine}. A larger distance indicates greater dissimilarity from existing papers, suggesting a higher likelihood of novelty.
(2) \textit{Contemporary Dissimilarity (CD)}: The average Euclidean distance between the generated abstract embedding and embeddings of the top 5 most similar abstracts in $B_{\text{con}}$. A smaller distance indicates greater similarity to newer papers, also suggesting a higher likelihood of novelty. 
(3) \textit{Contemporary Impact (CI)}: The average citation count of the top 5 most similar abstracts in $B_{\text{con}}$~\citep{yang2022gender}. A higher citation count suggests that the generated abstract is more likely to have a higher impact. 
To ensure comparability, we normalize each calculated metric using the mean value taken over all papers published in the same year as the chosen similar abstract in the corresponding database, with normalization defined as the metric divided by the mean value~\citep{alshebli2018preeminence}.

Since novelty is difficult to measure directly, we introduce a proxy metric to account for the three indicators: (4) \textit{Overall Novelty (ON)}. It is natural to assume that ON is positively related to both HD and CI and negatively related to CD, calculated as $\text{ON} = (\text{HD} \times \text{CI})/{\text{CD}}$. Mathematically, the expected value of ON is proportional to the novelty.

\begin{table}[t]
\centering
\begin{threeparttable}
\resizebox{\linewidth}{!}{
\begin{tabular}{c|cccccc}
\toprule
  \multirow{2}{*}{Method}   & \multirow{2}{*}{LLM$\uparrow$} & \multicolumn{2}{c}{Proposed Metric} & \multicolumn{3}{c}{Human Evaluation}\\
 &  & {CD$\downarrow$} & {CI$\uparrow$} & Nov$\uparrow$ & Fea$\uparrow$ & Eff$\uparrow$\\
\midrule
\multicolumn{7}{l}{\textit{Agent Model: GPT-4o}}\\
 HypoGen & 3.02 & \underline{0.36} & 3.10 & 4.78 & \underline{4.24} & 4.43 \\
AI Scientist  & \underline{3.10} & 0.38 & \underline{3.22} & \underline{4.94} & 4.18 & \underline{4.77}\\
 % \cellcolor{gray!15}{Ours} & \cellcolor{gray!15} 3.34 (\textcolor{blue}{+0.24})  & \cellcolor{gray!15} 0.34 (\textcolor{blue}{-0.02})  & \cellcolor{gray!15} 3.78 (\textcolor{blue}{+0.58}) & \cellcolor{gray!15} 5.24 (\textcolor{blue}{+0.30})  & \cellcolor{gray!15} 4.52 (\textcolor{blue}{+0.28})\\
  \cellcolor{gray!15}{\textbf{Ours}} & \cellcolor{gray!15} \textbf{3.34}  & \cellcolor{gray!15} \textbf{0.34}   & \cellcolor{gray!15} \textbf{3.78}  & \cellcolor{gray!15} \textbf{5.24}   & \cellcolor{gray!15} \textbf{4.52} & \cellcolor{gray!15} \textbf{4.95}\\
\multicolumn{7}{l}{\textit{Agent Model: LLaMA3.1-8b}}\\
 HypoGen & \underline{2.17} & 0.51 & \underline{2.16} & 3.54 & \underline{3.56} & 3.45\\
 AI Scientist & 2.09 & \underline{0.49} & 2.12 & \underline{3.66} & 3.52 & \underline{3.63} \\
 % \cellcolor{gray!15}{Ours} & \cellcolor{gray!15} 2.31 (\textcolor{blue}{+0.14}) &  \cellcolor{gray!15} 0.42 (\textcolor{blue}{-0.07})  & \cellcolor{gray!15} 3.29 (\textcolor{blue}{+1.13})  &  \cellcolor{gray!15} 4.08 (\textcolor{blue}{+0.42})  & \cellcolor{gray!15} 3.74 (\textcolor{blue}{+0.18})  \\
  \cellcolor{gray!15}{\textbf{Ours}} & \cellcolor{gray!15} \textbf{2.31} &  \cellcolor{gray!15} \textbf{0.42}   & \cellcolor{gray!15} \textbf{3.29}   &  \cellcolor{gray!15} \textbf{4.08}   & \cellcolor{gray!15} \textbf{3.74}  &
  \cellcolor{gray!15} \textbf{3.69} \\
 \multicolumn{7}{l}{\textit{Agent Model: LLaMA3.1-70b}}\\
 HypoGen & 2.18 & 0.49 & \underline{2.13} & 3.57 & \underline{3.61} & 3.52 \\
 AI Scientist & \underline{2.24} &  \underline{0.48}  & 2.11 & \underline{3.88} & 3.60 & \underline{3.66}\\
  % \cellcolor{gray!15}{Ours} & \cellcolor{gray!15} 2.53 (\textcolor{blue}{+0.29}) &   \cellcolor{gray!15} 0.40 (\textcolor{blue}{-0.08})  & \cellcolor{gray!15} 3.36 (\textcolor{blue}{+1.23})  &   \cellcolor{gray!15} 4.18 (\textcolor{blue}{+0.30})  & \cellcolor{gray!15} 3.84 (\textcolor{blue}{+0.23}) \\
    \cellcolor{gray!15}{\textbf{Ours}} & \cellcolor{gray!15} \textbf{2.53}  &   \cellcolor{gray!15} \textbf{0.40}   & \cellcolor{gray!15} \textbf{3.36}   &   \cellcolor{gray!15} \textbf{4.18}  & \cellcolor{gray!15} \textbf{3.84} & 
    \cellcolor{gray!15} \textbf{3.75} \\
\bottomrule
\end{tabular}}
\end{threeparttable}
\caption{Comparison with HypoGen and AI Scientist. The results demonstrate that our multi-agent system surpasses baseline models across all metrics. Here, ``Nov'', ``Fea'', and ``Eff'' denote human evaluation for ``Novelty'', ``Feasibility'', and ``Effectiveness'', respectively.}
\label{fig:comparisons_w_ai_scientist}
\end{table}

\noindent \textbf{Baselines.}~In this section, we compare \modelname\ with two recent agent-based systems for scientific discovery: the multi-agent system HypoGen~\cite{qi2024large} and the SOTA single-agent system AI Scientist~\cite{lu2024ai}. Given the differences in their pipelines, we adjust the experimental settings to a comparable level for a fair evaluation, elaborated in Appx.~\ref{sec:detail_comparison}. We then assess the results using our proposed metrics, LLM review scores, and human evaluations (Appx.~\ref{sec:human_evaluation}) to demonstrate the superiority of our method in idea generation.

\subsection{Analysis of Proposed Metrics} 
\label{sec:analyze_metrics}
\begin{figure}[ht]  
    \centering  
    \includegraphics[width=0.97\linewidth]{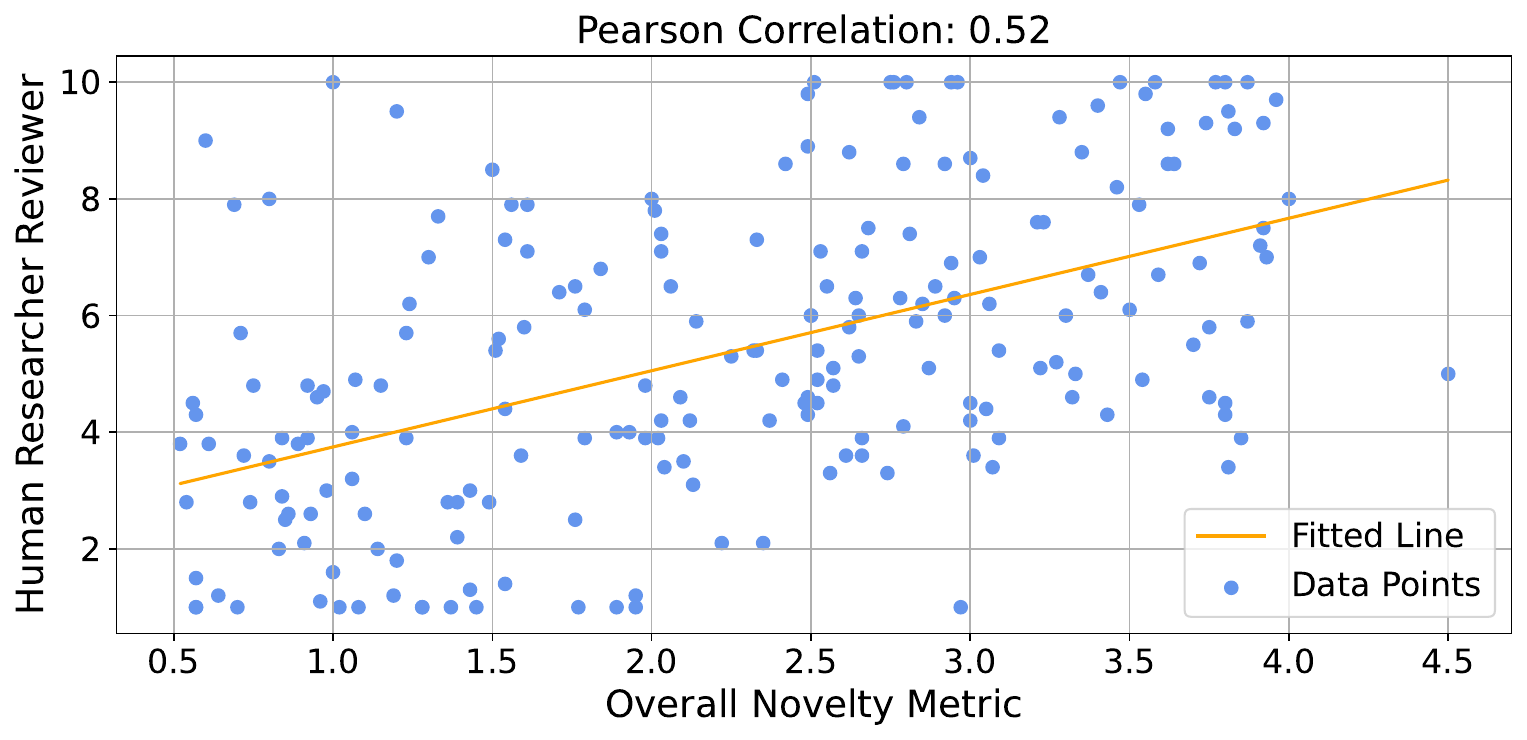}  
    \caption{Evaluation of abstracts using our overall novelty metric and human evaluation. The Pearson correlation coefficient of 0.52 indicates a positive correlation.}
    \label{fig:human_metric_main}
\end{figure}

To assess the validity of our proposed overall novelty (ON) metric, we selected 200 abstracts generated from the Computer Science dataset, which were evaluated using both ON and by human experts in the field of computer science (details in Appx.~\ref{sec:human_evaluation}). For the human evaluations, we the novelty scoring framework from~\citep{si2024can} as a guideline. To ensure objectivity, independent researchers assessed each abstract, and the average score was calculated.
As shown in Fig.~\ref{fig:human_metric_main}, the Pearson correlation coefficient between ON and human ratings demonstrates a positive correlation, which supports the validity of our metric. We also explore the correlation between ON and LLM-based review (a common novelty measurement method~\citep{gu2024survey}) in Appx.~\ref{app_metric} and conduct further analysis.

\subsection{Comparisons with Baselines} 
\label{sec:ai_scientist}
As shown in Tab.~\ref{fig:comparisons_w_ai_scientist}, our multi-agent system outperforms baseline models across metrics: our proposed metrics (CD, CI), LLM review, and human evaluation (Novelty, Feasibility, and Effectiveness), demonstrating its effectiveness in enhancing abstract novelty through collaboration. Notably, GPT-4o-based agents consistently achieve the highest novelty scores, reflecting its superior idea generation capability. In contrast, LLaMA3.1-8b and LLaMA3.1-70b-based agents show similar novelty scores, suggesting that moderate model size changes may not improve novelty.

% \subsection{Exploring SciSci: Team Dynamics}
\subsection{Exploring Collaboration Mechanism}
While the dynamics of group collaboration have been extensively studied in the Science of Science, its applicability to artificial multi-agent systems and the effects it may produce remain unclear. In this section, we analyze factors influencing idea novelty in our system, including team size, freshness, and research diversity—factors previously analyzed in human teams~\citep{wu2019large,zeng2021fresh,shi2023surprising}.
% \textcolor{red}{While the dynamics of group collaboration have been extensively studied in the Science of Science, its applicability to artificial multi-agent systems and the effects it may produce remain unclear. In this section, we analyze factors influencing idea novelty in our system, including team size, freshness, and research diversity—factors previously analyzed in human teams~\citep{wu2019large,zeng2021fresh,shi2023surprising}.}

\begin{figure}[t]  
    \centering  
    \begin{subfigure}[b]{\linewidth}
        \centering
        \includegraphics[width=\textwidth]{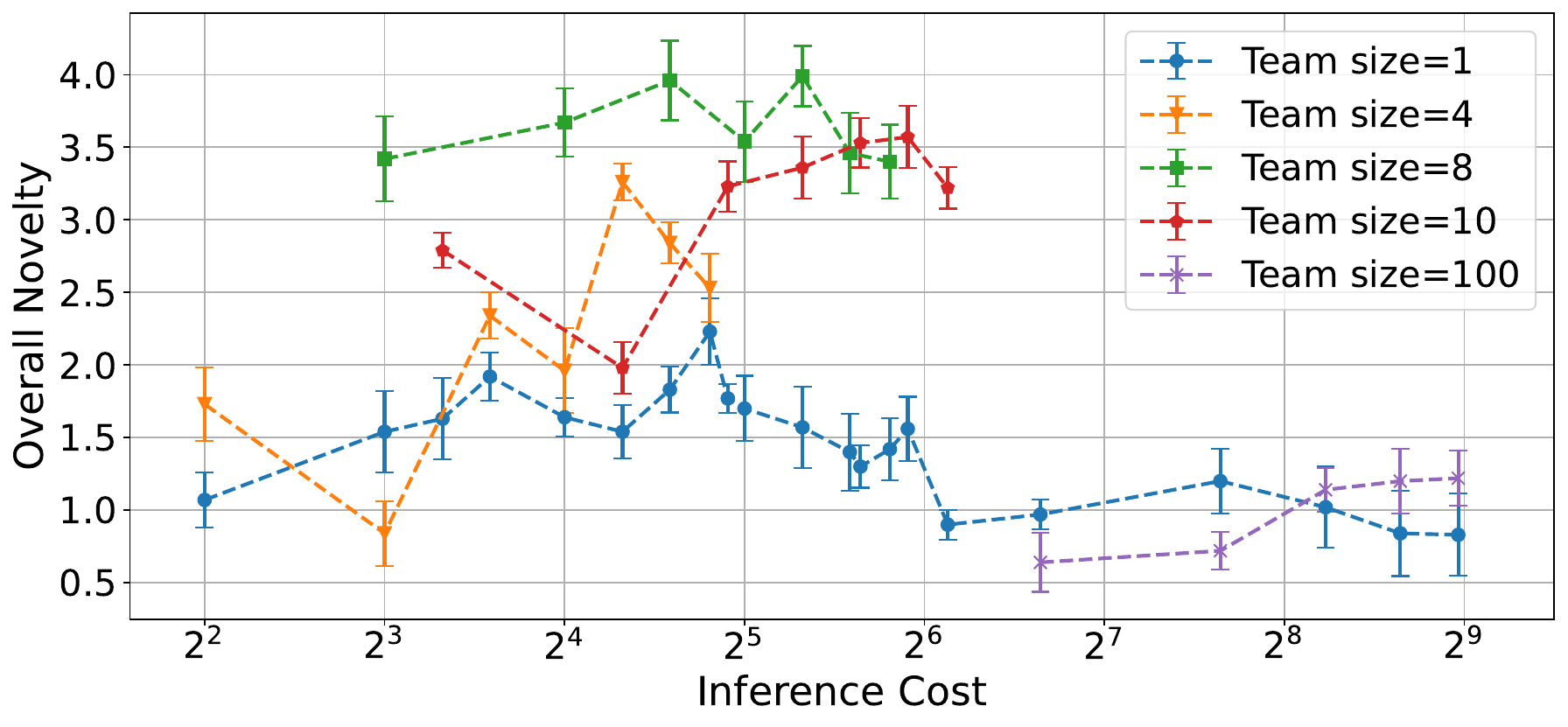}
        \caption{Experiments on the Computer Science dataset.}
    \end{subfigure}
    \hfill
    \centering
    \begin{subfigure}[b]{\linewidth}
        \centering
        \includegraphics[width=\textwidth]{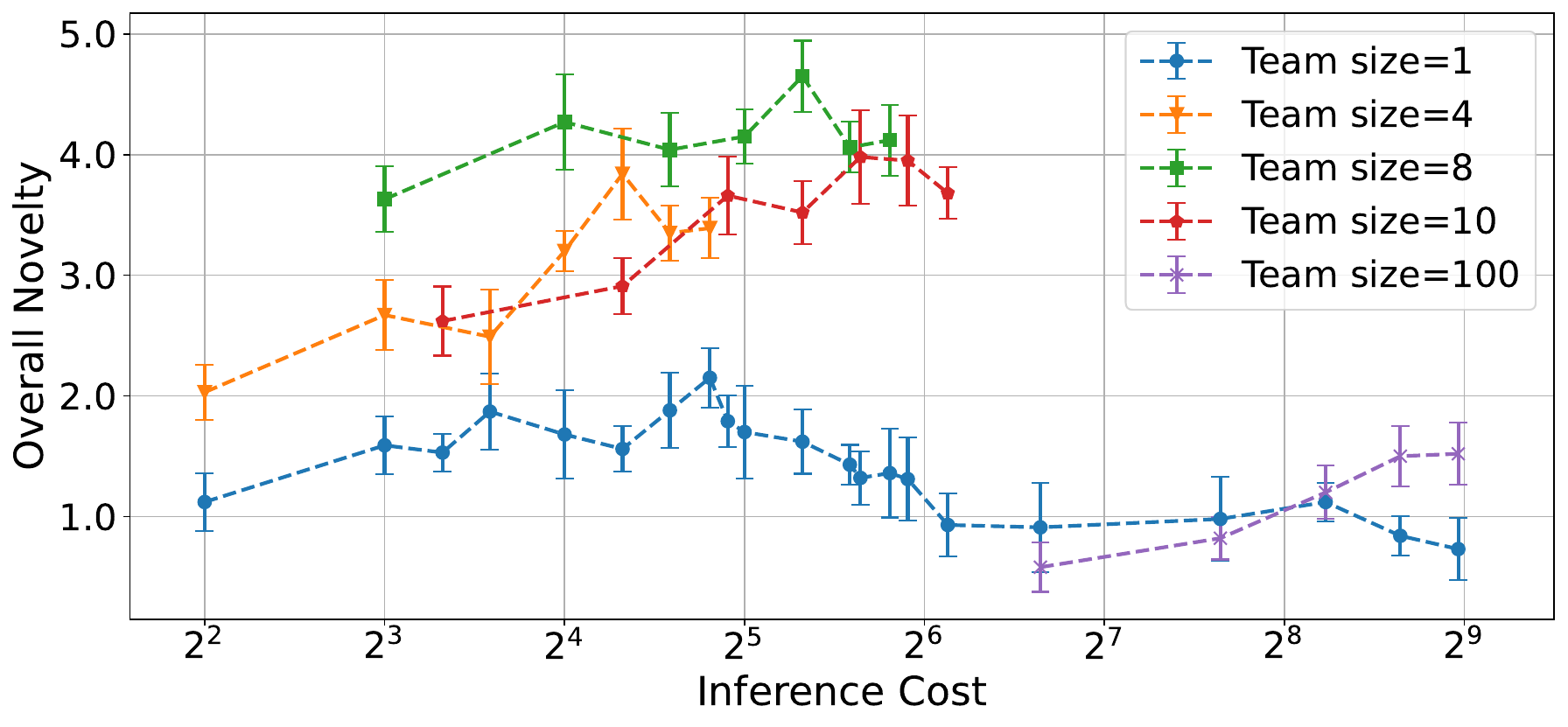}
        \caption{Experiments on the Open Academic Graph dataset.}
    \end{subfigure}
    \caption{Effects of team size and discussion turn on novelty. Peak occurs with 8 members and 5 turns, while larger teams or excessive turns hinder creativity. ``Inference Cost'' is the product of team size and turns.}
    \label{fig:team_size}
\end{figure}

\begin{figure}[t]  
    \centering  
    \begin{subfigure}[b]{\linewidth}
        \centering
        \includegraphics[width=\textwidth]{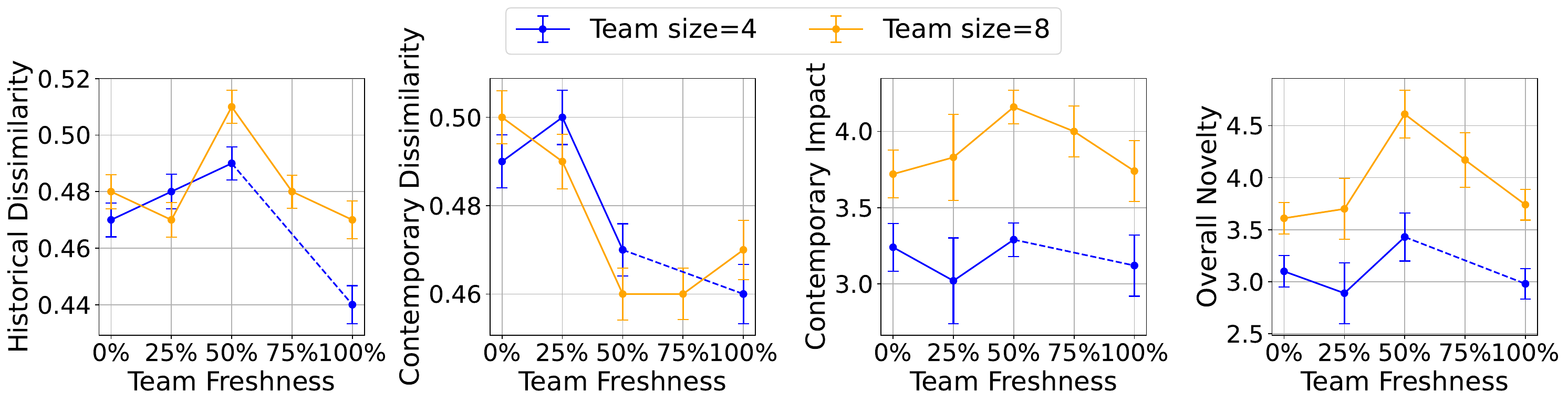}
        \caption{Experiments on the Computer Science dataset.}
    \end{subfigure}
    \hfill
    \begin{subfigure}[b]{\linewidth}
        \centering
        \includegraphics[width=\textwidth]{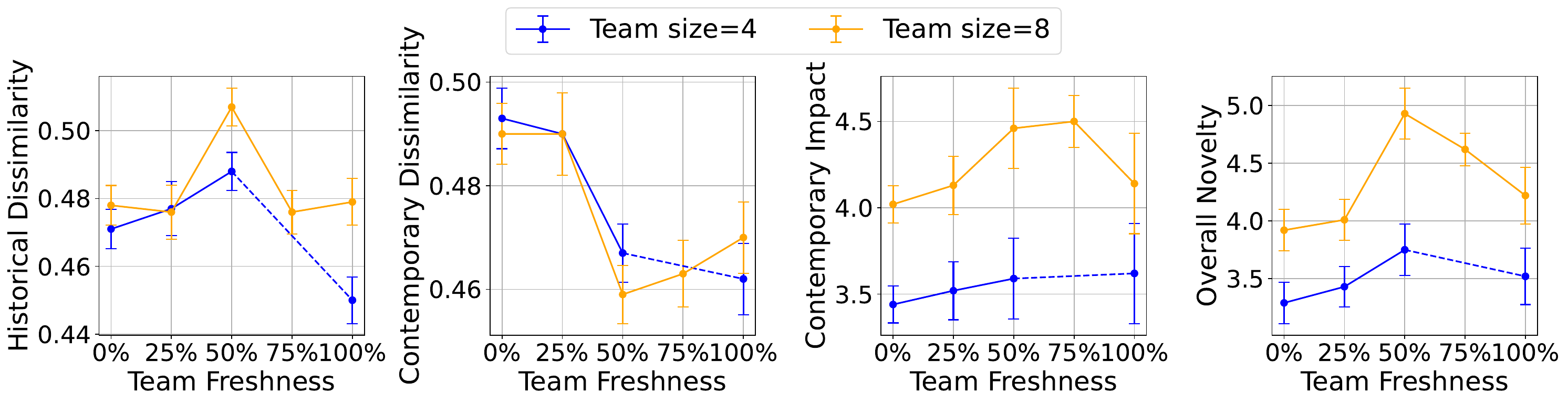}
        \caption{Experiments on the Open Academic Graph dataset.}
    \end{subfigure}
    \caption{The balance of new and returning collaborators in the team has a notable impact on novelty, with 50\% freshness yielding the highest historical dissimilarity and overall novelty, particularly in larger teams.}
    \label{fig:team_freshness}
\end{figure}

\noindent \textbf{Effects of Team Size on Novelty.}~The results in Fig.~\ref{fig:team_size} show that increasing team size can enhance the ON of generated abstracts by bringing in diverse ideas and perspectives. However, this relationship is not strictly linear, with peak ON occurring at a team size of 8. While moderate team expansion boosts novelty, excessively large teams can create coordination and communication challenges, leading to diluted contributions and groupthink. This aligns with existing literature, which suggests that smaller teams generate more disruptive ideas, while larger teams focus on refining existing concepts\citep{wuchty2007increasing, fortunato2018science, wu2019large}.

\noindent \textbf{Effects of Discussion Turn on Novelty.}~
The number of discussion turns plays a crucial role in fostering novel ideas~\citep{mezirow2003critical, li2023camel, shinn2024reflexion, lu2024ai}. As shown in our experimental results (Fig.~\ref{fig:team_size}), an optimal number of turns promotes deeper insights and more innovative research outputs. While the initial turns facilitate idea generation, excessive turns can result in fatigue and reduced engagement, ultimately hindering creativity. Our findings suggest that peak ON is achieved at around 5 discussion turns.
% The number of discussion turns significantly impacts novelty~\citep{mezirow2003critical, li2023camel, shinn2024reflexion, lu2024ai}. Experimental results (Fig.~\ref{fig:team_size}) show that an appropriate number of turns allows team members to explore, iterate, and refine ideas, fostering deeper understanding and more innovative research outputs. While initial turns aid idea generation, too many turns can lead to fatigue and diminished engagement, stifling creativity, where peak novelty is achieved with around 5 discussion turns.

\begin{figure}[t]  
    \centering  
    \begin{subfigure}[b]{\linewidth}
        \centering
        \includegraphics[width=\textwidth]{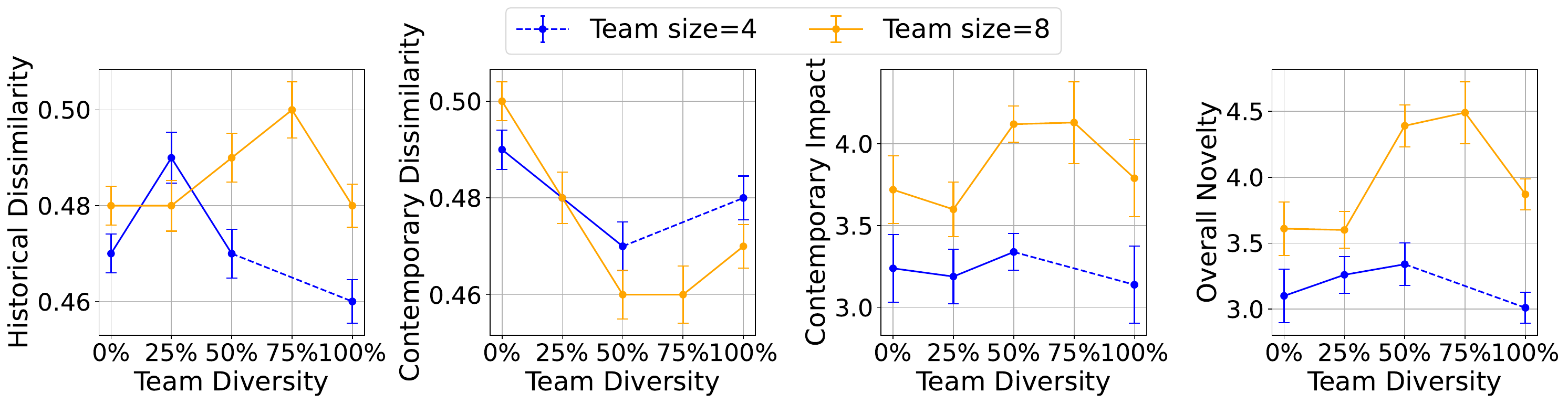}
        \caption{Experiments on the Computer Science dataset.}
    \end{subfigure}
    \hfill
    \begin{subfigure}[b]{\linewidth}
        \centering
        \includegraphics[width=\textwidth]{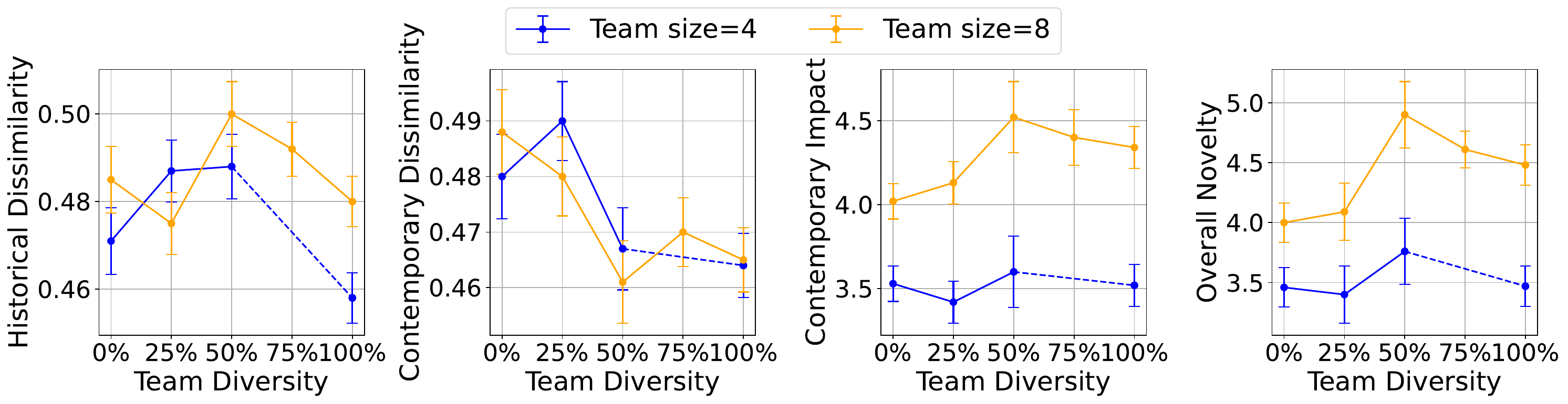}
        \caption{Experiments on the Open Academic Graph dataset.}
    \end{subfigure}
    \caption{Effects of team diversity on novelty. The optimal diversity level appears to be 50\% or 75\%, which maximizes novelty and impact across team sizes.}
    \label{fig:team_diversity}
\end{figure}

\noindent \textbf{Effects of Team Freshness on Novelty.}\label{sec:team_freshness}
As shown in Fig.~\ref{fig:team_freshness}, team freshness—the fraction of members who have not previously collaborated—affects the novelty of outputs. Freshness has its strongest effect at 50\%, especially for larger teams (size 8), where HD peaks. This suggests that a balanced mix of new and returning members promotes innovation by diverging from past research. As freshness increases, CD decreases, indicating that teams with more fresh members align their ideas more closely with future research trends. Both CI and ON reach their highest values around 50\% freshness, highlighting that a balanced team optimally combines novelty and future relevance, consistent with previous work~\citep{guimera2005team, zeng2021fresh}.

\noindent \textbf{Effects of Team Research Diversity on Novelty.} \label{sec:team_diversity} 
Team research diversity is defined as the proportion of team members specializing in distinct research topics. As shown in Fig.~\ref{fig:team_diversity}, moderate diversity boosts HD, with peak performance at 25-50\% diversity for 4-member teams and 50-75\% for 8-member teams across both datasets. CD drops before reaching 50\% diversity, suggesting that diverse teams align better with emerging research trends while maintaining innovation. Larger teams benefit more from higher diversity, showing an increase in CI, while smaller teams exhibit more stable, moderate effects. Overall, ON follows a reverse U-shaped curve for both team sizes and datasets, underscoring the importance of balancing research diversity for novel outcomes. This conclusion mirrors findings in Science of Science, where unexpected team combinations can enhance research impact~\citep{uzzi2013atypical, shi2023surprising}.

\section{Conclusion}
We introduce \modelname, an LLM-based multi-agent system that simulates the collaborative dynamics of scientific discovery. Our model focuses on the idea generation phase, demonstrating how specialized agents collaborate to generate diverse insights, reflecting real-world scientific teamwork. Experiments show that our method outperforms existing approaches in generating novel ideas and offer insights into the factors influencing collaborative research mechanisms.

\section{Limitations}
While our system effectively models scientific collaboration, it simplifies the complexities of real-world teamwork. In large research communities, multiple teams often work on related projects either collaboratively or independently, and researchers frequently participate in multiple teams simultaneously. Our current approach does not fully capture these intricate dynamics, potentially limiting its ability to model real-world scientific collaboration with high fidelity.

Beyond structural limitations, our system also inherits biases from the underlying language model and training data. If not properly mitigated, these biases could disproportionately favor well-established research domains while marginalizing emerging or underrepresented areas. This could reinforce existing disparities in scientific knowledge production, inadvertently shaping the trajectory of research based on historical patterns rather than fostering novel discoveries. Future work should explore methods to detect and counteract these biases, such as dataset diversification and bias-aware fine-tuning~\cite{ansar2025bias}.

Another critical limitation is the potential for generating plausible but incorrect or misleading scientific claims. This risk is particularly concerning in high-stakes fields such as medicine and policy-related research, where inaccuracies could lead to misinformation, flawed decision-making, or ethical concerns. Implementing stricter validation mechanisms, such as integrating fact-checking modules or leveraging expert human reviewers, will be necessary to ensure that generated content aligns with established scientific knowledge.

Additionally, our system could be misused for unethical purposes, such as automating the creation of fraudulent research papers or facilitating plagiarism. The ability to generate text that mimics academic writing raises concerns about the proliferation of low-quality or deceptive publications. To mitigate these risks, future work should explore safeguards such as watermarking AI-generated text~\cite{zhao2023provable}, developing robust detection algorithms for synthetic research content, and establishing ethical guidelines for responsible use.

Finally, the biases embedded in large-scale language model training data may disproportionately impact specific research communities. Researchers from underrepresented regions, institutions, or disciplines may find their work underemphasized or misrepresented, reinforcing systemic inequities in academia. Addressing these challenges requires continuous monitoring of dataset composition, proactive inclusion of diverse research sources, and collaboration with domain experts to refine training methodologies.

Our future research will also aim to scale up \modelname\ to a societal level, incorporating more dynamic and concurrent team interactions. Allowing agents to engage in multiple projects and collaborate across different teams would improve the realism of our simulations and better reflect the complexity of modern scientific collaboration. These advancements would also strengthen the system’s utility for the Science of Science community, enabling deeper investigations into how collaboration influences innovation and knowledge production.

\section{Ethics Statement}\label{ethics}
This research uses two publicly available datasets provided by AMiner, ensuring compliance with data privacy policies. Our system is designed to augment, not replace, human researchers, highlighting the need for human oversight to maintain the quality and integrity of generated outputs. 
We comply with the licensing terms of the LLMs used, as specified in their official terms of service. 
To promote research transparency, we have shared all relevant codes for reproducibility within the research community.

While our system supports scientific discovery, it carries potential risks. Data privacy concerns may arise even with public datasets if data handling is improper. To mitigate this, we have applied data anonymization to the datasets, such as masking real scientists’ names in simulations, to prevent data leakage and protect privacy. Additionally, stringent protocols have been implemented to minimize unintended discussions and potential misuse.

\section*{Acknowledgement}
This work is supported by Shanghai Artificial Intelligence Laboratory.

\bibliography{custom}

\clearpage

\appendix
\renewcommand{\contentsname}{Appendix}
\tableofcontents
\addtocontents{toc}{\protect\setcounter{tocdepth}{3}} % 设置目录深度为 1，即只包含章节级别

\begin{table*}[t]
\centering
\begin{threeparttable}
\begin{tabular}{cc|ccccccc}
\toprule
 \multirow{2}{*}{Future Database} & \multirow{2}{*}{Years}   & \multicolumn{7}{c}{Turns} \\ \cmidrule{3-9}
 &   & 1 & 2 & 3 & 4 & 5 & 6 & 7\\
\midrule
 OAG 3.1 & 2021-2023 & 1.75 & 0.90 & 2.36 & 2.00 &  3.40 &  2.91 & 2.53 \\
OAG 3.2 & 2024-2025 & 1.72 & 0.91 & 2.31 & 1.94 & 3.35 & 2.94 & 2.49 \\
\bottomrule
\end{tabular}
\end{threeparttable}
\caption{Comparison results on Open Academic Graph 3.1 and 3.2. The evaluation metric is ON.}
\label{tab:data_leakage}
\end{table*}

\section{Effect of the Potential Data Leakage} \label{appx:data_leakage}
We acknowledge that the use of papers published before 2024 may raise concerns about data leakage, given that the LLMs employed in our experiments are trained on data within this time period. However, this potential issue does not pose a significant threat to the validity of our experiments for the following reasons.
First, both the comparisons between our multi-agent system and baseline models, as well as the comparisons between different team settings, utilize the same LLMs. Since all models encounter the same exposure to training data from this period, any potential data leakage would affect all experiments equally. Thus, the relative performance differences we observe are not skewed by uneven data leakage. This ensures that the evaluation process remains fair and that the corresponding conclusions drawn are valid.
Moreover, our goal is not to demonstrate an absolute measure of novelty but rather to explore how different collaboration strategies and team settings influence the novelty of generated research outputs. As all team settings face the same potential exposure to historical data, the novelty metrics still provide an accurate comparison of the agents' ability to generate distinct and original ideas under varying conditions. Third, although LLMs are familiar with certain well-known academic papers and theoretical concepts, our extensive testing has shown that they cannot accurately replicate the abstract of the original work. Therefore, since our evaluation focuses on the abstract, this limitation has minimal impact on our assessment.

To directly assess the impact of temporal data leakage, we conducted an additional experiment. Using Open Academic Graph 3.2~\footnote{\url{https://open.aminer.cn/open/article?id=67aaf63af4cbd12984b6a5f0}}, we tested a temporal isolation scenario where papers from 2010–2020 were retained as the Past Paper Database, while 30,000 papers from 2024–2025 were used as the new Contemporary Paper Database. Given that LLaMA 3.1's training data extends only until December 2023, this setup ensured that no "future" papers beyond the model’s training scope were accessible. The experiments were conducted with a team size of 4 and LLaMA3.1-8b, and citation counts were normalized according to the method described in our paper to remove the influence of publication years.

As illustrated in Table~\ref{tab:data_leakage}, when shifting the Contemporary Paper Database from 2021–2023 to 2024–2025, the novelty scores exhibited only a slight decrease, with values remaining consistent across different turns. This suggests that knowledge leakage from future papers does not significantly impact our conclusions.

In summary, while data leakage is a valid concern, it affects all models and settings uniformly in our experiments. Therefore, it does not undermine the relative comparisons we make or the conclusions we draw regarding collaboration strategies and team performance. 

\section{Methodological Details}
\subsection{Self-review}\label{sec:self_review}
A self-review mechanism is considered after $R_{abstract}$ is finalized to pre-check its novelty. In this self-review, the optimized abstract $R_{abstract}$ is provided to the team leader to assess novelty by comparing it to similar papers in $B_{past}$. The prompt is:
\begin{equation}
    Q_{review}=\langle Q_{check}, R_{abstract}, B_{past}(R_{abstract})\rangle
\end{equation}

If this is the first time undergoing the self-review and the team leader determines that the similarity to existing papers is too high, the abstract will undergo further revision. The evaluation $R_{review}$ will then be added to Eq.~(\ref{origin_abstract}) for the next revision round: 
\begin{equation}
\begin{split}
    Q_{1,0}=\langle Q_{abstract}, Q_{judgement}, R_{review}, \\B_{past}(R_{abstract}),R_{abstract}\rangle
\end{split}
\end{equation}

If the abstract undergoes a second self-review and still does not meet the novelty requirement, it will be discarded, and the team will generate a new idea. Once the self-review yields satisfactory results, the final abstract will be produced, and the system will terminate. However, this self-review mechanism introduces uncertainty in total inference cost, making it difficult to ensure fair experimental comparisons. We discuss the effectiveness of this module only in the ablation study (see Appx.~\ref{sec:component}).

\section{Experimental Settings}
\label{appx:experimental_settings}
\subsection{Implementation}
We implement our system on top of the Agentscope framework~\citep{agentscope}, which serves for LLM-empowered multi-agent applications. We evaluate our system using different publicly available LLMs: GPT-4o~\citep{openai}, LLaMA3.1-8b~\citep{llama3}, and LLaMA3.1-70b. GPT-4o is accessible exclusively via a public API, while the LLaMA3.1 models are open-weight and invoked using the Ollama~\citep{ollama_github} in our experiments. LLaMA3.1-8b is chosen as the default LLM considering both efficiency and capability, where more experiments using different LLMs are shown in Appx. \ref{app_models}. Each experimental run on LLaMA3.1-8b takes approximately 10 minutes on 1 NVIDIA A100 40G GPU within a team discussion setting of 4 members and 5 turns ($K$ = 5). All experimental results are averaged on 20 runs.
\subsection{Dataset Details}
{\subsubsection{Computer Science Dataset.}~We first build our scientific research ecosystem using real scientists' information from the AMiner Computer Science dataset~\footnote{\url{https://www.aminer.cn/aminernetwork}}, which was constructed by extracting scientists' profiles from online web databases~\citep{tang2008arnetminer}. This dataset includes 1,712,433 authors and 2,092,356 papers, covering the period from 1948 to 2014, with disambiguated author names. To manage the large volume of data, we set $y_{start}$, $y_{bound}$, and $y_{end}$ to 2000, 2010, and 2014, respectively. 
For quality assurance, we filtered out past papers lacking abstracts or with fewer than 10 citations, contemporary papers with fewer than 5 citations or missing abstracts, and authors with fewer than 50 papers or 50 co-authors. As a result, we extracted detailed information from 156 authors and 178,197 papers to construct the ecosystem (85,217 papers for the past database and 92,980 papers for the contemporary database) and initialize the corresponding agents for the simulation. All paper and author data are embedded using the ``mxbai-embed-large'' model~\citep{emb2024mxbai}.}

{\subsubsection{Cross-domain Dataset.}~To demonstrate the generalizability and robustness of the system, we conduct additional experiments on the Open Academic Graph 3.1~\footnote{\url{https://open.aminer.cn/open/article?id=65bf053091c938e5025a31e2}}, which developed from the Open Academic Graph~\citep{zhang2022oag}.
This dataset includes 35,774,510 authors and 130,710,733 papers as of 2023, spanning diverse domains such as physics, chemistry, computer science, and biology. To manage this extensive dataset, we set $y_{start}$, $y_{bound}$, and $y_{end}$ to 2010, 2020, and 2023, respectively.
For quality control, we removed past papers without abstracts or with fewer than 200 citations, contemporary papers with fewer than 50 citations or missing abstracts, and authors with fewer than 40 papers or 50 co-authors. This filtering resulted in detailed information on 3,169 authors and 201,131 papers to construct the ecosystem (139,646 papers for the past database and 61,485 papers for the contemporary database) and initialize agents for simulation.
Unlike the Computer Science dataset, the Open Academic Graph lacks author-level information on research interests, citations, and co-authors. To address this, we extracted each author’s published papers between $y_{start}$ and $y_{bound}$ and enriched their profiles with the keywords, citations, and authors associated with these papers. Given that the complete set of keywords from all an author’s published papers could be overwhelming, we prompted GPT-4o~\citep{openai} to extract and summarize these keywords into concise research interests for each author. All paper and author data were similarly embedded using the “mxbai-embed-large” model.}
\subsection{Detailed Comparison Settings}\label{sec:detail_comparison}
Given the problem formulation of the AI Scientist, HypoGen, and ours, we must make several justifications to ensure relative fairness in our comparisons:
(1) Since the AI Scientist is limited to generating ideas from predefined topics (2D Diffusion, NanoGPT, and Grokking), we include NanoGPT in the topic selection prompt for both HypoGen and \modelname\ as the initial discussion topic, ensuring that the final abstracts align with the same research direction.
(2) Given the different approaches to idea generation, we ensure that comparisons are made under similar inference costs. The AI Scientist performs 50 turns of self-reflection during its idea generation, which does not apply to its paper (abstract) generation.
To match the inference costs, we set \modelname\ with 4 team members and 5 discussion turns, and set HypoGen to 12 discussion turns, ensuring that the experiments are conducted under comparable computational costs.
(3) Since the AI Scientist and HypoGen lack a scientific research ecosystem, they retrieve papers across all time ranges through the Semantic Scholar API~\citep{fricke2018semantic} or PubMed\footnote{\url{https://pubmed.ncbi.nlm.nih.gov/}}. To maintain consistency, we replace our databases and HypoGen's PubMed API with the Semantic Scholar API for paper retrieval and metric calculation. 
Specifically, after generating ideas and corresponding abstracts, we use the generated ideas as queries to retrieve related papers, extracting the relevant abstracts and citation counts for evaluation.
(4) We evaluate the generated abstracts using both our metrics (CD and CI), the AI Scientist's metric (LLM review score)~\citep{lu2024ai} and the human evaluation~\citep{si2024can}. The LLM review score is calculated by GPT-4o, which conducts abstract reviews based on a truncated version of the Neural Information Processing Systems (NeurIPS) conference review guidelines\footnote{\url{https://neurips.cc/Conferences/2024/ReviewerGuidelines}}, shown in Fig.~\ref{fig:llm_review}.

\begin{table}[ht]
\centering
\begin{threeparttable}
\resizebox{\linewidth}{!}{
\begin{tabular}{cc|c}
\toprule
 \multicolumn{2}{c|}{Database}   & \multirow{2}{*}{ON $\uparrow$} \\ 
 Idea Generation  & Novelty Assessment  & \\ \midrule
- & - & 2.60  \\
\scalebox{1.3}{\ding{51}} & - & 3.62  \\ 
 - & \scalebox{1.3}{\ding{51}} & 2.76 \\
 \rowcolor{gray!15}
 \scalebox{1.3}{\ding{51}} & \scalebox{1.3}{\ding{51}} & 4.23  \\
\bottomrule
\end{tabular}}
\end{threeparttable}
\caption{Effects of the paper database in the processes of idea generation and novelty assessment on the Computer Science dataset.}
\label{tab:table_database}
\end{table}

\begin{table}[ht]
\centering
\begin{threeparttable}
\resizebox{\linewidth}{!}{
\begin{tabular}{cc|c}
\toprule
 \multicolumn{2}{c|}{Database}   & \multirow{2}{*}{ON $\uparrow$} \\ 
 Idea Generation  & Novelty Assessment  & \\ \midrule
- & - & 2.41   \\
\scalebox{1.3}{\ding{51}} & - &  3.99 \\ 
 - & \scalebox{1.3}{\ding{51}} & 2.60 \\
  \rowcolor{gray!15}
 \scalebox{1.3}{\ding{51}} & \scalebox{1.3}{\ding{51}} & 4.65  \\
\bottomrule
\end{tabular}}
\end{threeparttable}
\caption{Effects of the paper database in the processes of idea generation and novelty assessment on the Open Academic Graph dataset.}
\label{tab:table_database_2}
\end{table}

\subsection{Human Evaluation}\label{sec:human_evaluation}
To comprehensively compare our method with baseline models, we additionally conduct a human evaluation to assess novelty, feasibility, and effectiveness besides our proposed metrics and LLM review score. Ten PhD students in computer science (the relevant field for the selected research topic), unaware of the method identities, rated the abstracts on a 10-point scale (1: Poor, 10: best) based on three metrics~\citep{si2024can}: 1. Novelty: Whether the idea is creative and different from existing works on the topic, and
brings fresh insights; 2. Feasibility: How feasible it is to implement and execute this idea as a research project; 3. Effectiveness: How likely the proposed idea is going to work well (e.g., better than existing baselines). The 10-point scale guideline is provided for human experts and is designed based on the principles outlined by~\citet{si2024can}.

\begin{table*}[t]
\centering
\begin{threeparttable}
\begin{tabular}{cc|ccccccc}
\toprule
 \multirow{2}{*}{Team Size} & \multirow{2}{*}{Invitation Mechanism}   & \multicolumn{7}{c}{Turns} \\ \cmidrule{3-9}
   &   & 1 & 2 & 3 & 4 & 5 & 6 & 7\\
\midrule
 & - & 1.67 & 0.75 & 2.29 & 1.92 &  3.30 &  2.78 & 2.47 \\
 \rowcolor{gray!15}
 \cellcolor{white}\multirow{-2}{*}{4} & \scalebox{1.3}{\ding{51}} & 1.75 & 0.90 & 2.36 & 2.00 &  3.40 &  2.91 & 2.53 \\ \midrule
 & - & 3.36 & 3.53 & 3.88 & 3.49 & 4.12 &  3.37 & 3.30 \\
 \rowcolor{gray!15}
 \cellcolor{white}\multirow{-2}{*}{8}& \scalebox{1.3}{\ding{51}} & 3.48 & 3.67 & 3.97 & 3.56 &  4.23 &  3.48 & 3.43 \\

\bottomrule
\end{tabular}
\end{threeparttable}
\caption{Effects of invitation mechanism in team discussion. Comparison results show that this mechanism helps to improve the novelty of outputs. The evaluation metric is ON.}
\label{tab:table1}
\end{table*}

\begin{table*}[ht]
\centering
\begin{threeparttable}
\begin{tabular}{cc|ccccccc}
\toprule
 \multirow{2}{*}{Team Size} & \multirow{2}{*}{Novelty Assessment}   & \multicolumn{7}{c}{Turns} \\ \cmidrule{3-9}
   &   & 1 & 2 & 3 & 4 & 5 & 6 & 7\\
\midrule
 & - & 1.56 & 0.73 & 2.20 & 1.87 &  3.19 &  2.74 & 2.39 \\
  \rowcolor{gray!15}
 \cellcolor{white}\multirow{-2}{*}{4} & \scalebox{1.3}{\ding{51}} & 1.75 & 0.90 & 2.36 & 2.00 &  3.40 &  2.91 & 2.53 \\ \midrule
 & - & 3.28 & 3.41 & 3.82 & 3.43 & 3.98 &  3.27 & 3.25 \\
   \rowcolor{gray!15}
 \cellcolor{white}\multirow{-2}{*}{8} & \scalebox{1.3}{\ding{51}} & 3.48 & 3.67 & 3.97 & 3.56 &  4.23 &  3.48 & 3.43 \\

\bottomrule
\end{tabular}
\end{threeparttable}
\caption{Effects of novelty assessment. Comparison results show that novelty assessment helps to improve the novelty of outputs. The evaluation metric is ON.}
\label{tab:table2}
\end{table*}

\begin{table*}[!ht]
\centering
\begin{threeparttable}
\begin{tabular}{cc|ccccccc}
\toprule
 \multirow{2}{*}{Team Size} & \multirow{2}{*}{Self-review}   & \multicolumn{7}{c}{Turns} \\ \cmidrule{3-9}
   &   & 1 & 2 & 3 & 4 & 5 & 6 & 7\\
\midrule
 & - & 1.60 & 0.74 & 2.26 & 1.89 &  3.26 &  2.77 & 2.41 \\
   \rowcolor{gray!15}
 \cellcolor{white}\multirow{-2}{*}{4}& \scalebox{1.3}{\ding{51}} & 1.75 & 0.90 & 2.36 & 2.00 &  3.40 &  2.91 & 2.53 \\ \midrule
 & - & 3.32 & 3.44 & 3.84 & 3.45 & 3.99 &  3.33 & 3.27 \\
   \rowcolor{gray!15}
 \cellcolor{white}\multirow{-2}{*}{8 }& \scalebox{1.3}{\ding{51}} & 3.48 & 3.67 & 3.97 & 3.56 &  4.23 &  3.48 & 3.43 \\
\bottomrule
\end{tabular}
\end{threeparttable}
\caption{Effects of self-review in abstract generation. Comparison results show that self-review after abstract generation helps to check and improve the novelty of outputs. The evaluation metric is ON.}
\label{tab:table3}
\end{table*}

\begin{table*}[ht]
\centering
\begin{threeparttable}
\begin{tabular}{cc|ccccccc}
\toprule
 \multirow{2}{*}{Team Size} & \multirow{2}{*}{Discussion Topology}   & \multicolumn{7}{c}{Turns} \\ \cmidrule{3-9}
   &   & 1 & 2 & 3 & 4 & 5 & 6 & 7\\
\midrule
 \rowcolor{gray!15}
 \cellcolor{white}\multirow{2}{*}{4} & Sequential Mode & 1.75 & 0.90 & 2.36 & 2.00 &  3.40 &  2.91 & 2.53 \\
 & Random Mode & 1.67 & 0.85 & 2.34 & 2.01 &  3.35 &  2.84 & 2.47 \\ \midrule
 \rowcolor{gray!15}
 \cellcolor{white}\multirow{2}{*}{8} & Sequential Mode & 3.48 & 3.67 & 3.97 & 3.56 & 4.23 &  3.48 & 3.43 \\
 & Random Mode & 3.50 & 3.63 & 3.88 & 3.52 &  4.16 &  3.40 & 3.39 \\

\bottomrule
\end{tabular}
\end{threeparttable}
\caption{Comparison between two discussion topologies: sequential mode and random mode. The results show that sequential mode generally outperforms random mode in most cases as it leverages more background knowledge.}
\label{tab:ablation_topology}
\end{table*}

\section{Ablation Study}

\subsection{Effects of Paper Database}\label{sec:ablation_database}

In the workflow of the proposed \modelname, the past paper database is utilized as a reference during idea generation and novelty assessment. To verify the role of the constructed paper database, we conduct ablation studies on both datasets under the setting of 8 team members and 5 discussion turns, as shown in Tab. \ref{tab:table_database} and \ref{tab:table_database_2}. The experimental results demonstrate that having references enhances the novelty of the final generated ideas, avoiding shallow or fanciful outcomes produced by the system.

\subsection{Effects of Components Designed for Improving Novelty}\label{sec:component}
In this section, we respectively test the effects of the invitation mechanism in team discussion, the role of the novelty assessment step, and the impact of self-review in abstract generation. All experiments are conducted with a 5-turn discussion. The results consistently show improvements in ON when these components are applied.  
\subsubsection{Invitation Mechanism}
For the invitation mechanism (results are presented in Tab.~\ref{tab:table1}), introducing new scientists into the discussion positively impacts the team's performance across both 4-member and 8-member teams. This indicates that seeking external insights from relevant but non-team scientists fosters more diverse and novel ideas.
\subsubsection{Novelty Assessment}
The novelty assessment step (results are presented in Tab.~\ref{tab:table2}) also influences the scores. If novelty assessment is not considered, then the output of idea generation will not be an idea list but the idea from the last scientist. Novelty assessment ensures that the generated ideas are continuously evaluated for originality, helping teams avoid overlap with existing research. The improvement is more noticeable in larger teams, where more ideas are being generated and assessed.

\subsubsection{Self-review}
Finally, the self-review mechanism (results are presented in Tab.~\ref{tab:table3}) is crucial in further refining the abstracts. By allowing the team leader to re-evaluate the abstract for novelty after it is fully generated, low-quality abstracts are discarded, and the team engages in a new discussion to generate a better idea, as evidenced by the score improvements for both team sizes.

\begin{table*}[!ht]
\centering
\begin{threeparttable}
\begin{tabular}{cc|cccc|cc}
\toprule
 \multirow{2}{*}{Team Size}& \multirow{2}{*}{Pattern} & \multicolumn{4}{c|}{Turns}   & \multirow{2}{*}{Cost $\downarrow$} & \multirow{2}{*}{ON $\uparrow$} \\ \cmidrule{3-6}
 & & Topic & Idea & Check & Abstract & &\\
\midrule
 & Fixed  & 5 & 5 & 5 & 5 & 80 &3.26\\
  \rowcolor{gray!15}
 \cellcolor{white}\multirow{-2}{*}{4}  & Adaptive &  2.4  & 4.5  & 3.2 & 4.2 & \textbf{57.2} &\textbf{3.49}\\ \midrule
  & Fixed &  5 & 5   &  5 & 5 & 160 &3.99 \\
   \rowcolor{gray!15}
 \cellcolor{white}\multirow{-2}{*}{8} & Adaptive & 2.9  &  4.8  & 4.0 & 3.8 & \textbf{124.0} &\textbf{4.37}  \\
\bottomrule
\end{tabular}
\end{threeparttable}
\caption{Comparison between fixed turns and adaptive turns in team discussions. The time taken by collaborator selection and team discussion's invitation mechanism is not counted, and self-review is not employed for a better comparison. The adaptive pattern shows both a lower inference cost and a higher ON.}
\label{table:ablation_adaptive}
\end{table*}

\subsection{Effects of Discussion Pattern} 
\subsubsection{Discussion Topologies}
\label{sec:ablation_topologies}
\begin{figure}[ht]  
    \centering  
    \includegraphics[width=\linewidth]{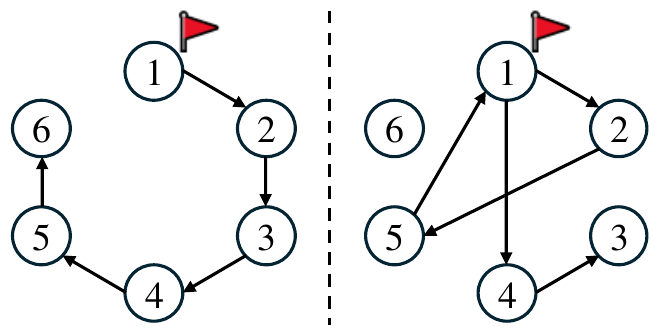}  
    \caption{Two different discussion topologies: sequential mode (\textit{Left}), and random mode (\textit{Right}). In sequential mode, participants take turns presenting their ideas in a round-table format, one at a time. In random mode, after an agent speaks, the next speaker is chosen randomly, with the restriction that the previous speaker cannot be selected consecutively. Note that the actual discussion topology of our system is more complex, benefiting from the proposed invitation mechanism.}
    \label{fig:topology}
\end{figure}

To further explore the effect of the discussion pattern, particularly discussion topologies, we compare two modes: sequential mode and random mode. As shown in Fig.~\ref{fig:topology}, in sequential mode, participants take turns presenting their ideas in a round-table format; in random mode, after an agent speaks, the next speaker is chosen randomly, with the restriction that the previous speaker cannot be selected consecutively. \textbf{Note that the actual discussion topology of our system is more complex, benefiting from the proposed invitation mechanism}. The comparison results are presented in Tab.~\ref{tab:ablation_topology}, with tests conducted for team sizes of 4 and 8. It can be observed that sequential mode generally outperforms random mode in most cases, as it leverages a broader knowledge base from the scientist agents. In random mode, some agents remain silent (i.e., are skipped) when the number of turns is limited. As a result, sequential mode is selected as the default discussion topology.

% \begin{table*}[ht]
% \centering
% \begin{threeparttable}
% \begin{tabular}{cc|ccccccc}
% \toprule
%  \multirow{2}{*}{Team Size} & \multirow{2}{*}{Distribution}   & \multicolumn{7}{c}{Turns} \\ \cmidrule{3-9}
%    &   & 1 & 2 & 3 & 4 & 5 & 6 & 7\\
% \midrule
% \multirow{2}{*}{4} & Uniform Distribution (Origin) & 1.75 & 0.90 & 2.36 & 2.00 &  3.40 &  2.91 & 2.53 \\
%  & Normal Distribution & 1.78 & 0.88 & 2.40 & 2.06 &  3.42 &  2.88 & 2.55 \\ \midrule
% \multirow{2}{*}{8} & Uniform Distribution (Origin) & 3.48 & 3.67 & 3.97 & 3.56 & 4.23 &  3.48 & 3.43 \\
%  & Normal Distribution & 3.51 & 3.70 & 4.03 & 3.55 &  4.27 &  3.50 & 3.42 \\

% \bottomrule
% \end{tabular}
% \end{threeparttable}
% \caption{Effect of the explore mechanism in scientific collaboration on the Computer Science dataset. The evaluation metric is ON.}
% \label{tab:increment}
% \end{table*}

\subsubsection{Discussion Turns}\label{sec:ablation_adaptive}
In the previous experiments, we fixed the number of discussion turns in each step to ensure fair comparisons. However, in real-world research environments, teams of scientists do not spend the same amount of time on each stage of the research process. To explore this, we compare fixed discussion turns with adaptive turn numbers. 
In the adaptive pattern, the team leader decides whether the team needs additional turns based on the current progress and the goals of each stage.
The results of both patterns, along with their corresponding inference cost, are shown in Tab.~\ref{table:ablation_adaptive}. The comparison reveals that the adaptive pattern achieves a higher ON while reducing the inference cost. This efficiency can be attributed to the more flexible approach, allowing teams to adjust their discussions dynamically rather than adhering to a rigid structure (which may lead the team in the wrong direction when a section is over-discussed).
Furthermore, examining the number of turns at each stage in both 4-person and 8-person teams under the adaptive pattern offers additional insights. Larger teams require more discussion turns and face greater challenges in reaching consensus~\citep{janis1972victims,pitters2014psychology}. This highlights the adaptive pattern's advantage in accommodating the complexities of larger teams while maintaining a higher level of novelty in the final research outputs.

\begin{figure}[ht]  
    \centering  
    \includegraphics[width=0.99\linewidth]{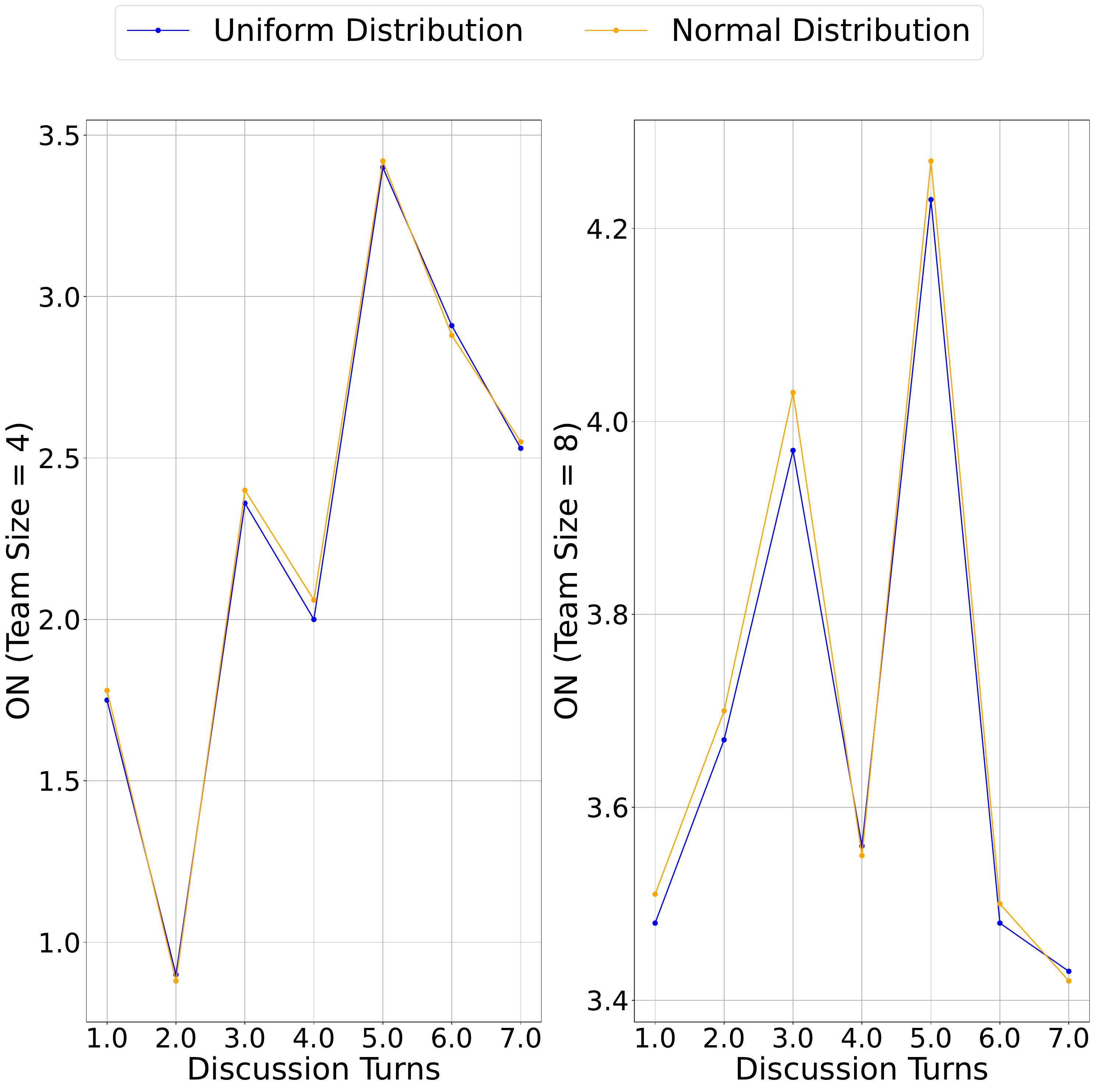}  
    \caption{Effect of the explore mechanism in scientific collaboration on the Computer Science dataset. Variations in the distribution of exploration strategies do not significantly affect the overall novelty of generated ideas.}
    \label{fig:increment}
\end{figure}

{
\subsection{Effects of Different Exploration Mechanisms on Scientific Collaboration}\label{app_initialization}
If we regard the collaboration among our scientists as an "explore-exploit" model, collaborating with scientists they have not worked with before can be viewed as ``explore", while collaborating with previously partnered scientists can be seen as ``exploit". In this section, we further investigate how the mechanism of "explore" impacts our final results.}

{
In the default setting, when creating the adjacency matrix, we add 1 to each value in the matrix to ensure that agents collaborate with individuals they have not worked with before, treating "explore" as uniformly distributed. Based on this, we also conducted experiments where the incremental values follow a normal distribution, defined as:
\begin{equation}\label{app_function}
    f(x)=\frac{1}{\sqrt{2\pi}} e^{-\frac{(x-\mu)^2}{2\sigma^{2}}},
\end{equation}
where $\mu=1$ and $\sigma=1$.
}

{
Next, we sample $x$ from this probability density function, using the absolute value $|x|$ as the increment in the adjacency matrix instead of a constant value of 1. The results of different explore mechanisms are presented in Fig. \ref{fig:increment}. The additional experiments demonstrate that replacing the original uniform distribution with a normal distribution has a positive impact, but the optimal distribution remains to be further explored. 
}

\begin{figure*}[ht]  
    \centering  
    \includegraphics[width=\linewidth]{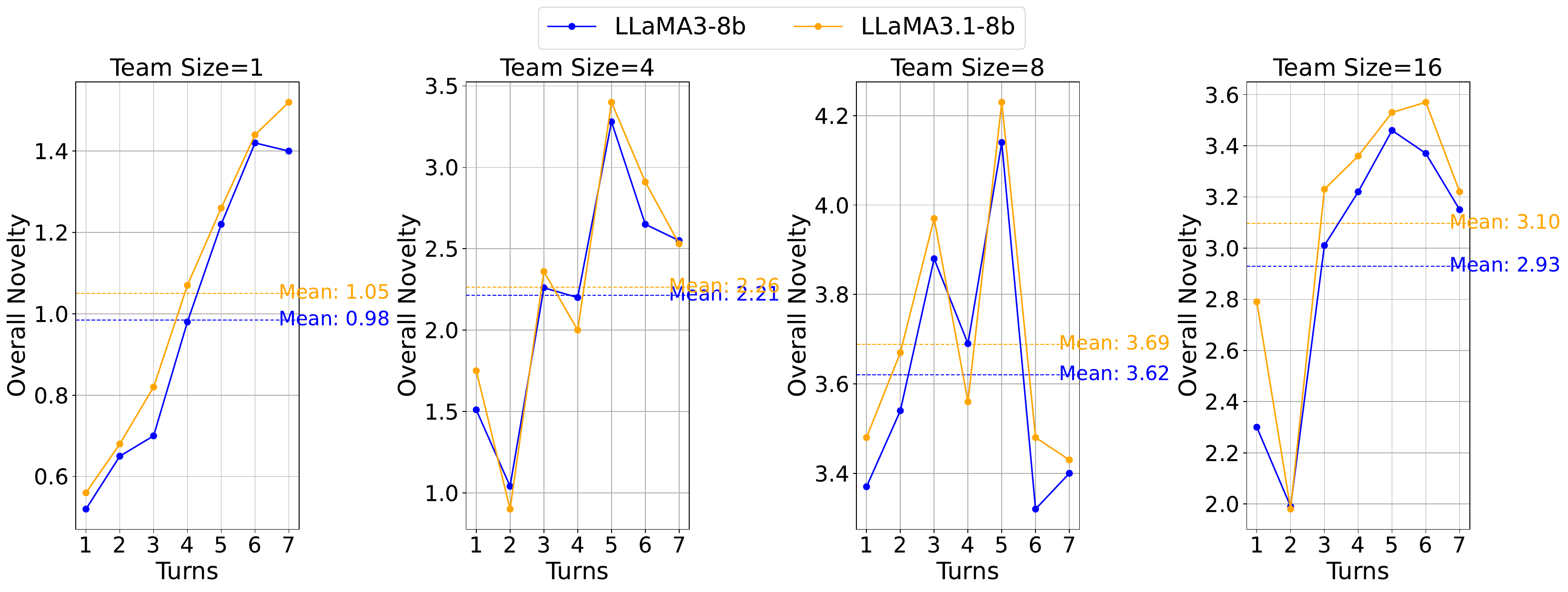}  
    \caption{The impact of team size and turn count on system performance for both LLaMA3-8b and LLaMA3.1-8b models on the Computer Science dataset. Despite variations in LLM capabilities, the conclusions about the influence of team size and discussion turn count remain consistent.}
    \label{fig:agent_model_comparison}
\end{figure*}

% \begin{table*}[ht]
% \centering
% \begin{threeparttable}
% \begin{tabular}{c|cccccccc}
% \toprule
%  \multirow{2}{*}{Team Size}   & \multicolumn{8}{c}{Turns} \\ \cmidrule{2-9}
%    &  1 & 2 & 3 & 4 & 5 & 6 & 7 & Mean\\
% \midrule
% 1  & 0.52 & 0.65 & 0.70 & 0.98 &  1.22 &  1.42 & 1.40 & 0.98\\
% 4  & 1.51 & 1.04 & 2.26 & 2.20 &  3.28 &  2.65 & 2.55 & 2.21 \\
%   \rowcolor{gray!15}
%  \cellcolor{white} {8}  & 3.37 & 3.54 & 3.88 & 3.69 & 4.14 &  3.32 & 3.40 & 3.62 \\
% 10  & 2.30 & 1.99 & 3.01 & 3.22 &  3.46 &  3.37 & 3.15 & 2.92\\

% \bottomrule
% \end{tabular}
% \end{threeparttable}
% \caption{The impact of team size and turn count on system performance across LLaMA3-8b on the Computer Science dataset. The evaluation metric is ON.}
% \label{tab:llama3-8b}
% \end{table*}

% \begin{table*}[ht]
% \centering
% \begin{threeparttable}
% \begin{tabular}{c|cccccccc}
% \toprule
%  \multirow{2}{*}{Team Size}   & \multicolumn{8}{c}{Turns} \\ \cmidrule{2-9}
%    &  1 & 2 & 3 & 4 & 5 & 6 & 7 & Mean\\
% \midrule
% 1  & 0.56 & 0.68 & 0.82 & 1.07 &  1.26 &  1.44 & 1.52 &  1.05\\
% 4  & 1.75 & 0.90 & 2.36 & 2.00 &  3.40 &  2.91 & 2.53 & 2.26\\
%   \rowcolor{gray!15}
%  \cellcolor{white} {8}  & 3.48 & 3.67 & 3.97 & 3.56 & 4.23 &  3.48 & 3.43 & 3.69\\
% 10  & 2.79 & 1.98 & 3.23 & 3.36 &  3.53 &  3.57 & 3.22 & 3.10\\

% \bottomrule
% \end{tabular}
% \end{threeparttable}
% \caption{The impact of team size and turn count on system performance across LLaMA3.1-8b on the Computer Science dataset. The evaluation metric is ON.}
% \label{tab:llama3.1-8b}
% \end{table*}

{
\subsection{Effects of Different Underlying LLMs}\label{app_models}
Given the potential variation in capabilities across LLMs, the impact of team size and turn count on system performance may vary across different models. Therefore, we additionally incorporated the open-source model LLaMA3-8b as the underlying LLM, which is inferior to LLaMA3.1-8b. By comparing the two LLMs with significantly different capabilities, the pattern can be discovered.
}

{
The experimental results on both LLaMA3-8b and LLaMA3-8b are shown in Fig.~\ref{fig:agent_model_comparison}. It can be observed that using LLaMA3-8b results in overall performance that is inferior to LLaMA3.1-8b, due to the inherently weaker capabilities of LLaMA3-8b. Regarding the effects of team size on novelty, whether with LLaMA3 or LLaMA3.1, a moderate team size enhances novelty. While multi-agent teams outperform single agents, excessively large teams may face coordination challenges and communication barriers. For the effects of discussion turn on novelty, our analysis shows that an optimal number of turns allows team members to refine and explore better ideas. While the initial turns contribute to idea generation, an excessive number of turns can lead to fatigue and diminished engagement. Overall, despite variations in LLM capabilities, the conclusions about the influence of team size and discussion turn count remain consistent.
}

\section{More Analysis of Proposed Metrics}\label{app_metric}

In this paper, we propose objective evaluation metrics to assess the novelty of system outputs, primarily based on vector similarity. This approach aligns with the research in the Science of Science domain, where our computational methods (relying on vector union and overlap) draw on established literature, such as ~\citet{boyack2005mapping, shi2023surprising, liu2023data}.

Another potential concern is the introduction of bias due to a lack of diversity in the datasets or the generated ideas. To address this, we employ two large-scale datasets in our experiments to ensure content diversity. Additionally, each scenario in our experiments is tested 20 times to ensure that the conclusions regarding the generated ideas are not influenced by potential biases.

\begin{figure}[ht]  
    \centering  
    \includegraphics[width=\linewidth]{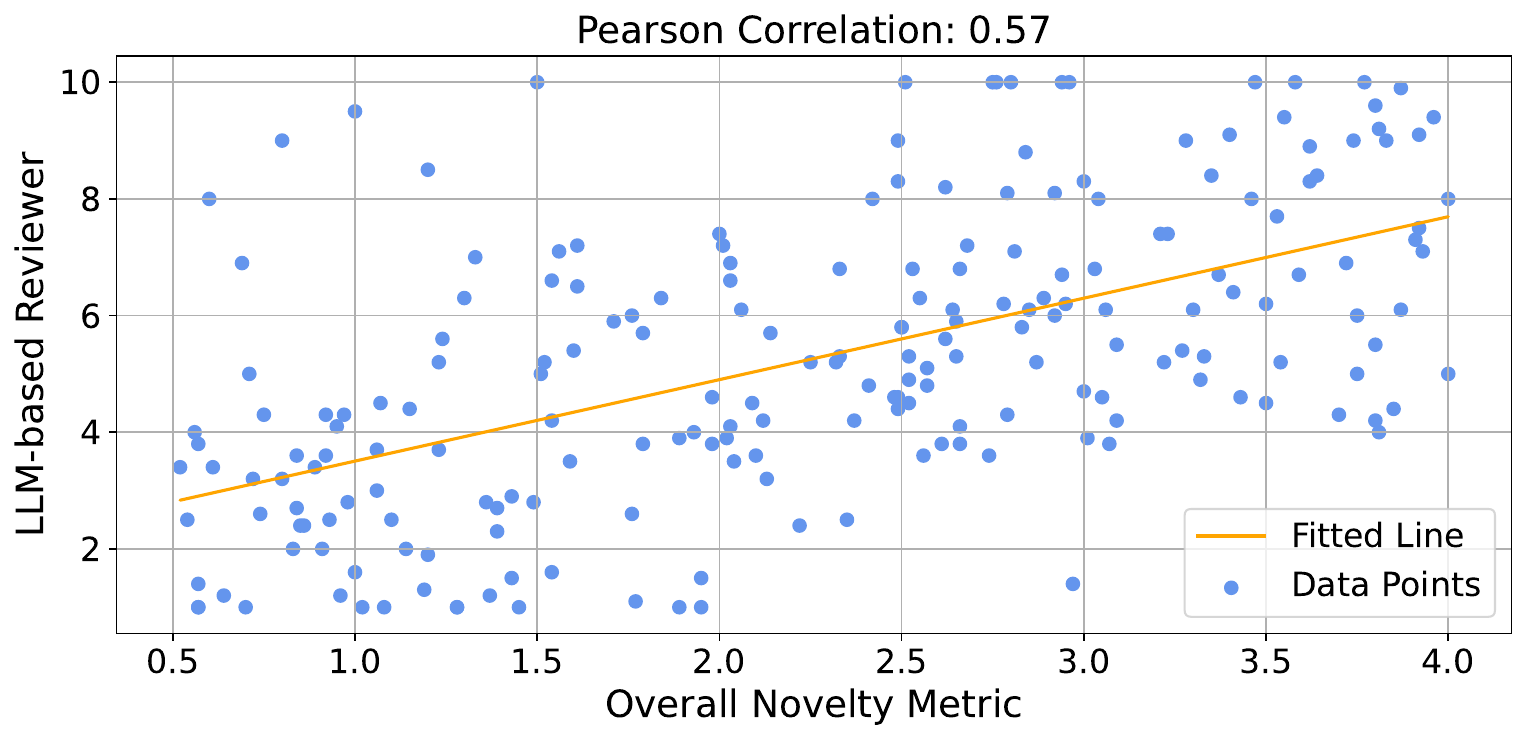}  
    \caption{The evaluation results of the same abstract under two different review metrics: our proposed overall novelty metric and LLM-based reviewer. The Pearson correlation coefficient equals 0.57, denoting the positive correlation of our metric with the LLM-based reviewer.}
    \label{fig:LLM_metric}
\end{figure}

\begin{figure}[ht]  
    \centering  
    \includegraphics[width=\linewidth]{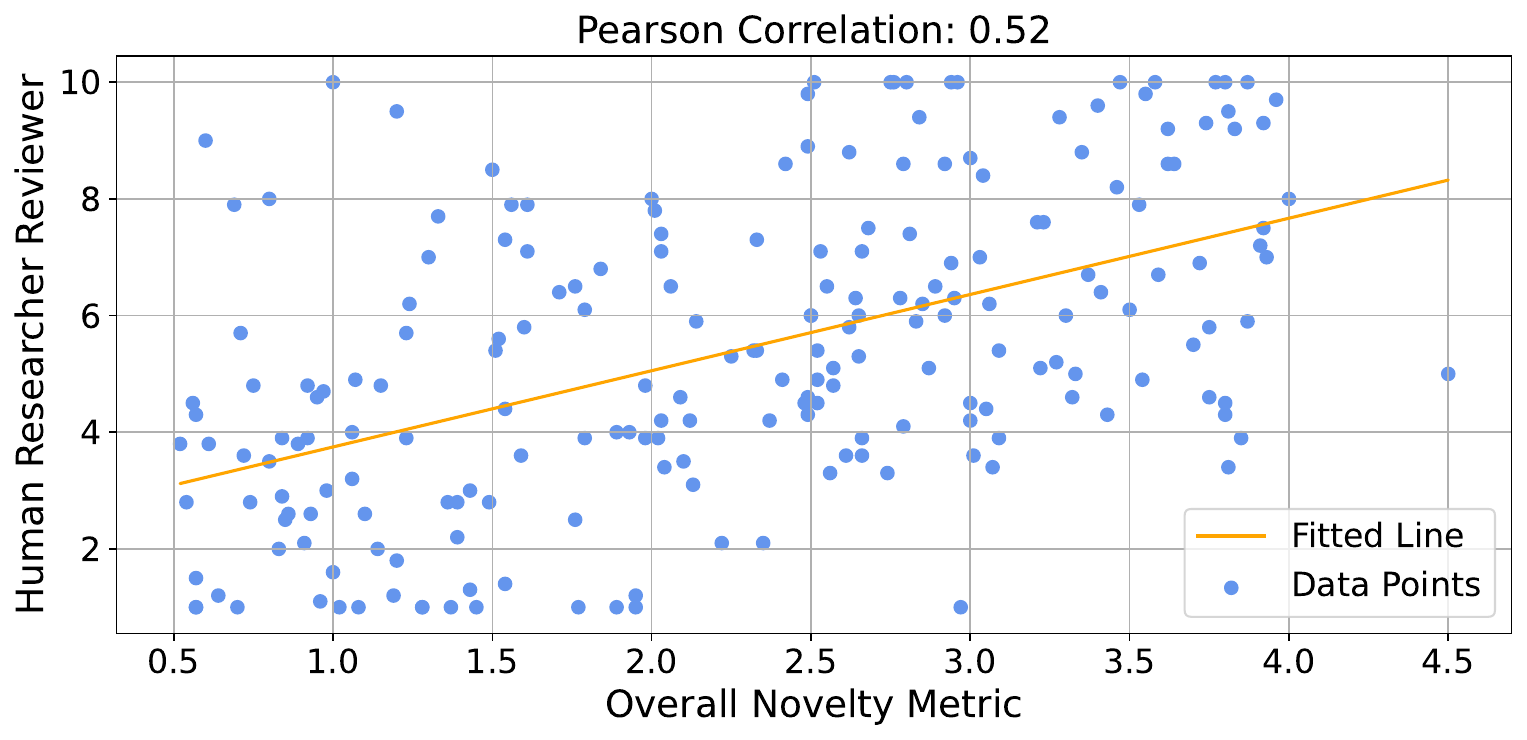}  
    \caption{The evaluation results of the same abstract under two different review metrics: our proposed overall novelty metric and human researcher. The Pearson correlation coefficient equals 0.52, denoting the positive correlation of our metric with the human researcher.}
    \label{fig:human_metric}
\end{figure}

To further assess the validity of our proposed overall novelty metric, we selected 200 abstracts generated under various experimental conditions from the Computer Science dataset. These abstracts were evaluated using three approaches: (1) our proposed overall novelty metric, (2) LLM-based reviewers (utilizing the GPT-4o API), and (3) human researchers specializing in computer science (detailed in Appx.~\ref{sec:human_evaluation}). For both LLM-based reviewers and human researchers, we employed the novelty scoring framework from~\citep{si2024can} as a guideline. To enhance objectivity, each abstract was evaluated three times by LLM-based reviewers, with the average score calculated. Similarly, three independent human researchers assessed each abstract, and their average score was computed.
The evaluation results are presented in Fig.~\ref{fig:LLM_metric} and Fig.~\ref{fig:human_metric}. In Fig.~\ref{fig:LLM_metric}, the axes represent the scores of the same abstract evaluated under our proposed metric and the LLM-based reviewers. In Fig.~\ref{fig:human_metric}, the axes represent the scores of the same abstract evaluated under our proposed metric and by human researchers. The Pearson correlation coefficients between our proposed overall novelty metric and both LLM-based reviewers and human researchers demonstrate a positive correlation with established novelty measurement methods~\citep{lu2024ai,si2024can}, which, to some extent, supports the validity of our metric.

\begin{figure*}[ht]  
    \centering  
    \includegraphics[width=\textwidth]{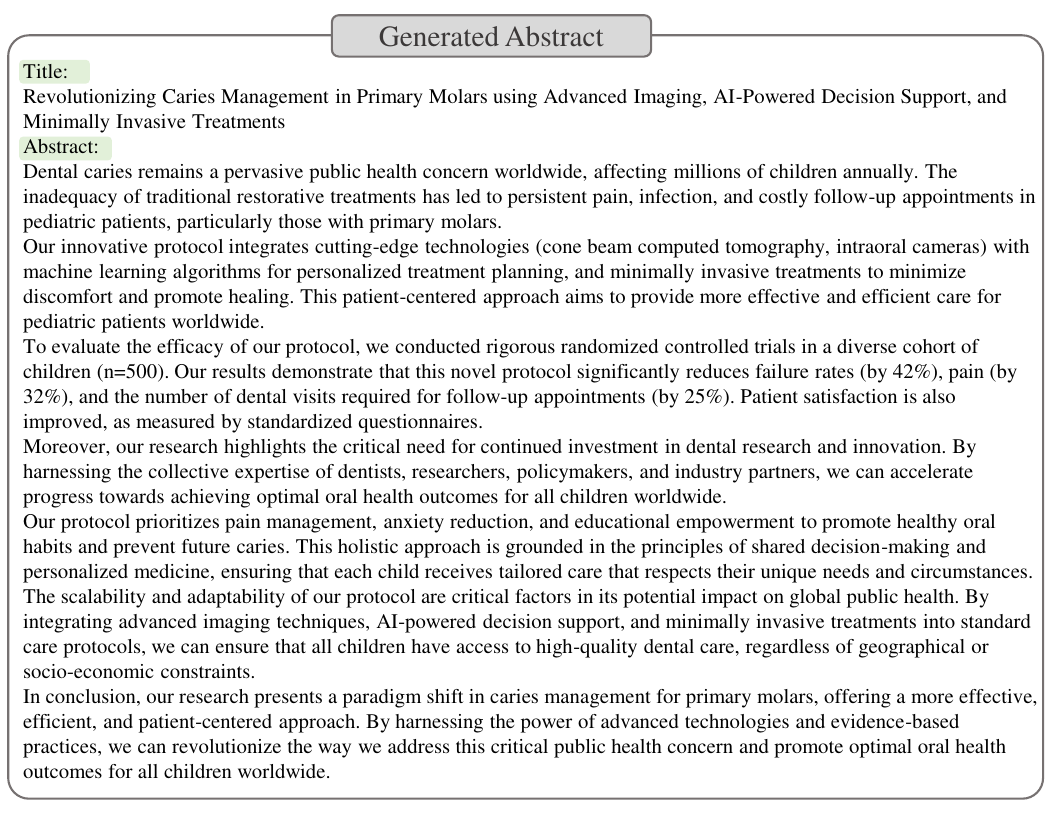}  
    \caption{The first example abstract generated by our \modelname. This abstract discusses the application of artificial intelligence in caries management.}
    \label{fig:generate_abstract_1}
\end{figure*}

\section{Discussions on System Feasibility}\label{sec:discussion_feasibility}
\begin{figure*}[ht]  
    \centering  
    \includegraphics[width=\textwidth]{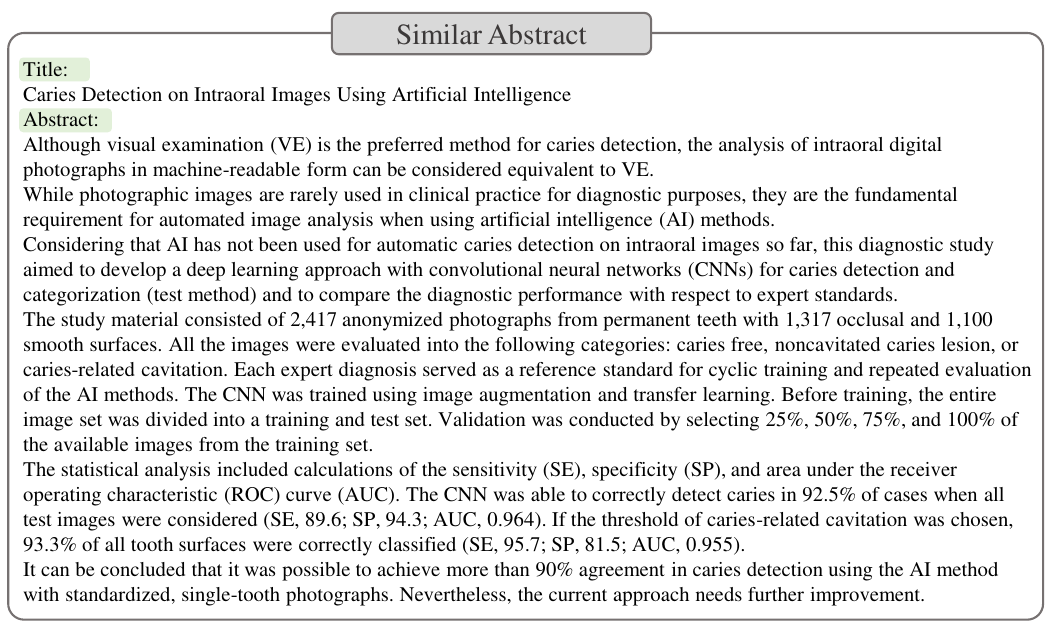}  
    \caption{The recently published similar paper~\citep{kuhnisch2022caries} of abstract, corresponding to the first example abstract generated by our \modelname. This abstract discusses the application of artificial intelligence in caries management.}
    \label{fig:similar_abstract_1}
\end{figure*}

\begin{figure*}[ht]  
    \centering  
    \includegraphics[width=\textwidth]{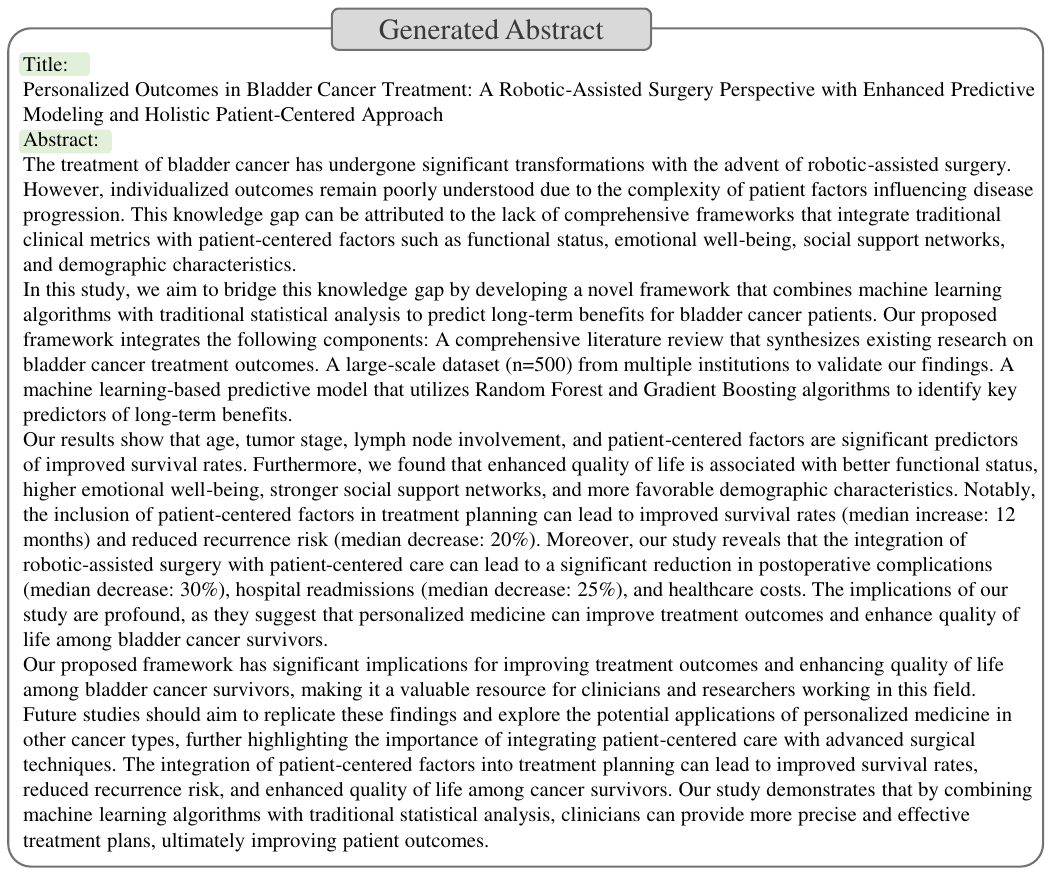}  
    \caption{The second example abstract generated by our \modelname. This abstract discusses the application of robots in cancer treatment.}
    \label{fig:generate_abstract_2}
\end{figure*}

\begin{figure*}[ht]  
    \centering  
    \includegraphics[width=\textwidth]{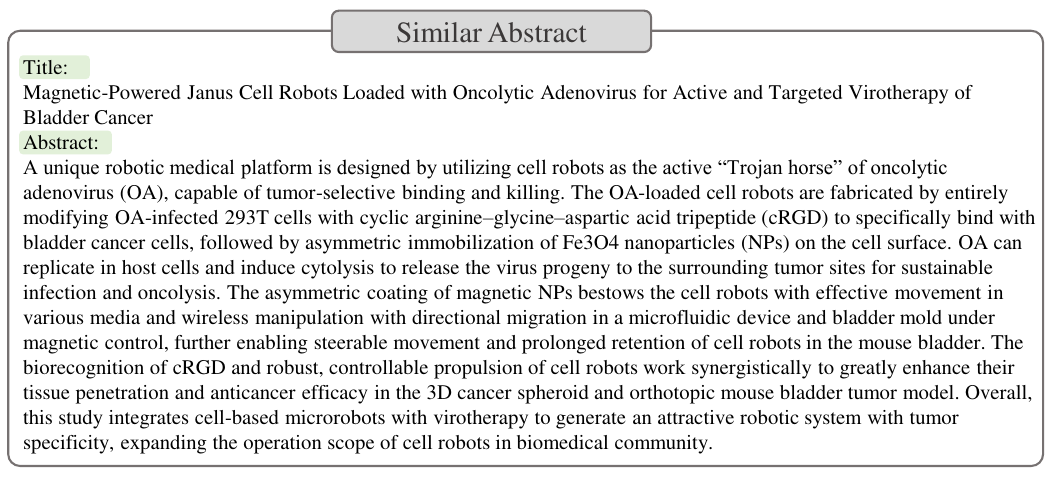}  
    \caption{The recently published similar paper~\citep{cong2022magnetic} of abstract, corresponding to the second example abstract generated by our \modelname. This abstract discusses the application of robots in cancer treatment.}
    \label{fig:similar_abstract_2}
\end{figure*}

To better show the feasibility of the proposed \modelname, we present two example abstracts generated by our system using the Open Academic Graph 3.1 dataset from 2011 to 2020, alongside corresponding similar recently published papers, to highlight the practical relevance and applicability of our system’s outputs. The first example abstract generated by our system is shown in Fig. \ref{fig:generate_abstract_1}, while the corresponding similar paper~\citep{kuhnisch2022caries} of the abstract is shown in Fig. \ref{fig:similar_abstract_1}. This pair of abstracts discusses the application of artificial intelligence in caries management. The second example abstract generated by our system is shown in Fig. \ref{fig:generate_abstract_2}, while the corresponding similar paper~\citep{cong2022magnetic} of the abstract is shown in Fig. \ref{fig:similar_abstract_2}. This pair of abstracts discusses the application of robots in cancer treatment. Overall, these examples demonstrate that our system has the potential to discover novel scientific ideas.

\section{Consistency Between Two Datasets.}~As shown in Fig.~\ref{fig:team_size}, no significant differences were found in outputs from single-agent teams (team size equals 1) across the two datasets, as individual research lacks interdisciplinary input. The overall trend—novelty rising, peaking at a team size of 8 and 5 turns, then declining—was consistent across both datasets, demonstrating our platform's robustness and the potential of multi-agent systems in enhancing scientific idea generation. These findings also support the value of interdisciplinary collaborations for driving higher-impact research~\citep{shi2023surprising}.

\section{Prompts}\label{sec:prompts}
\subsection{Scientist Definition}
\begin{figure*}[ht]  
    \centering  
    \includegraphics[width=\textwidth]{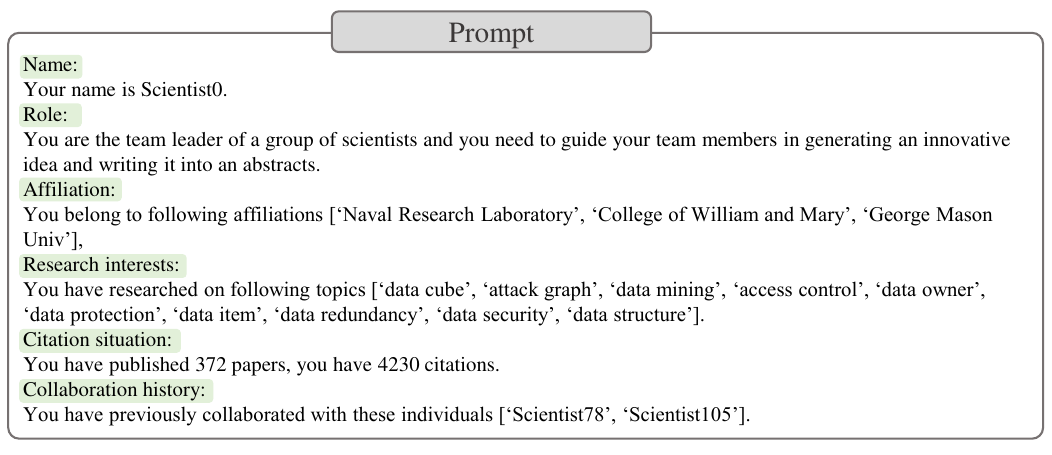}  
    \caption{The system prompt of each scientist agent is the personal information, including the name, role, affiliation, research interests, citation situation, and collaboration history.}
    \label{fig:scientist_prompt}
\end{figure*}

We use the personal information of the scientist to define the agent, where the corresponding system prompt is illustrated in Fig.~\ref{fig:scientist_prompt}.

\subsection{Collaboration Selection}
\begin{figure*}[ht]  
    \centering  
    \includegraphics[width=\textwidth]{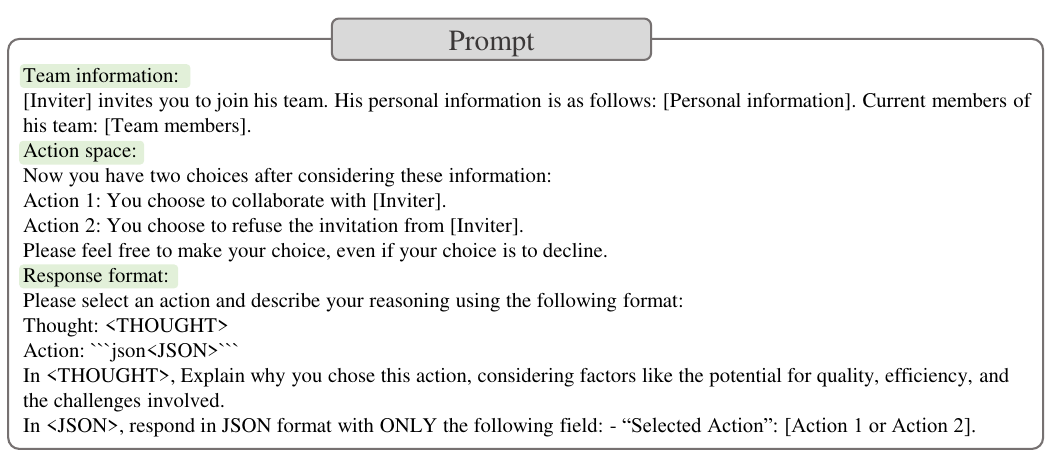}  
    \caption{The prompt for the collaboration selection.}
    \label{fig:collaboration_prompt}
\end{figure*}

The prompt for collaboration selection is illustrated in Fig.~\ref{fig:collaboration_prompt}.

\subsection{Topic Discussion}
\subsubsection{Discussion}
\begin{figure*}[ht]  
    \centering  
    \includegraphics[width=\textwidth]{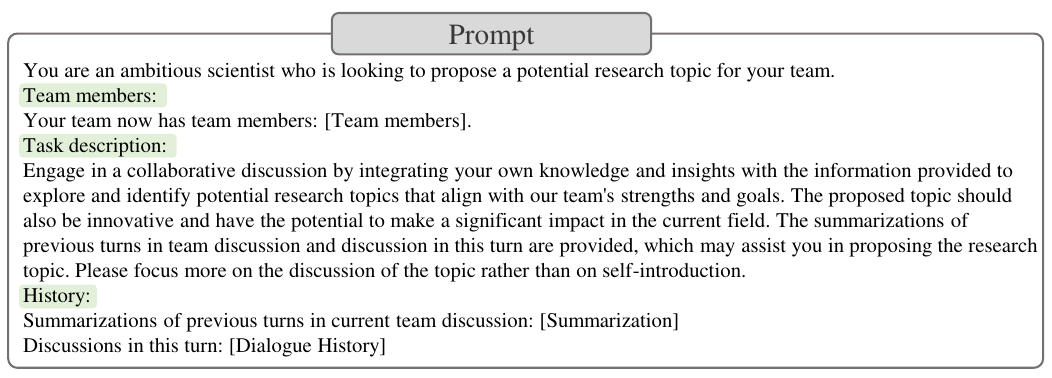}  
    \caption{The prompt for the topic discussion.}
    \label{fig:discuss_prompt}
\end{figure*}

The prompt for the topic discussion is illustrated in Fig.~\ref{fig:discuss_prompt}.

\subsubsection{Summarization}
\begin{figure*}[ht]  
    \centering  
    \includegraphics[width=\textwidth]{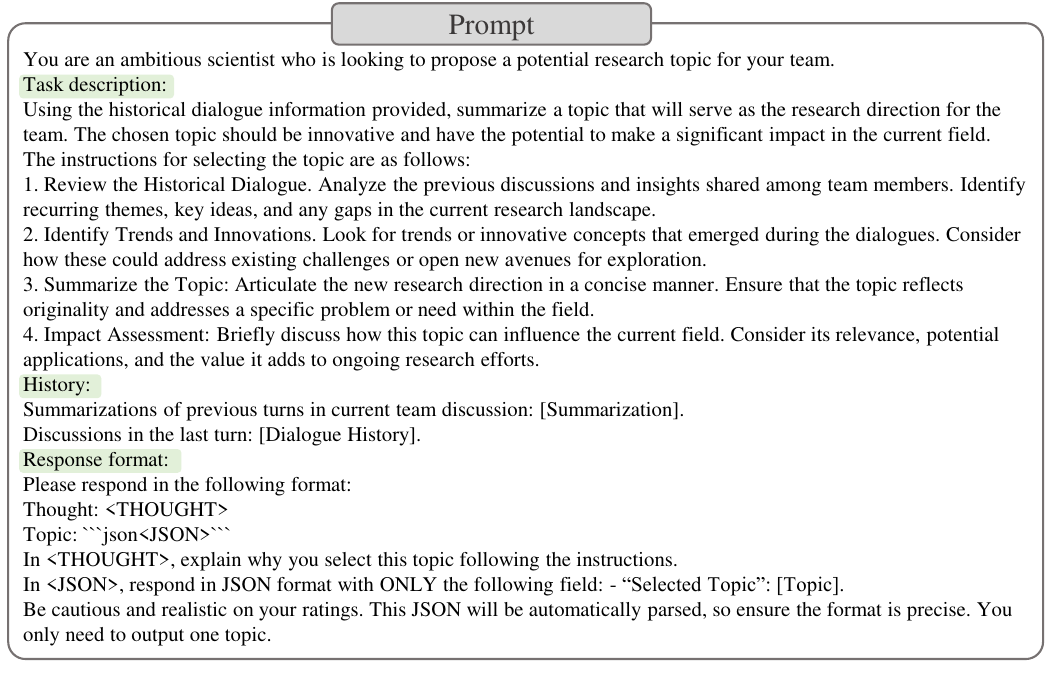}  
    \caption{The prompt for the final topic selection after topic discussion.}
    \label{fig:discuss_summary_prompt}
\end{figure*}

The prompt for the final topic selection after several turns of topic discussion is illustrated in Fig.~\ref{fig:discuss_summary_prompt}.

\subsection{Idea Generation}
\begin{figure*}[ht]  
    \centering  
    \includegraphics[width=\textwidth]{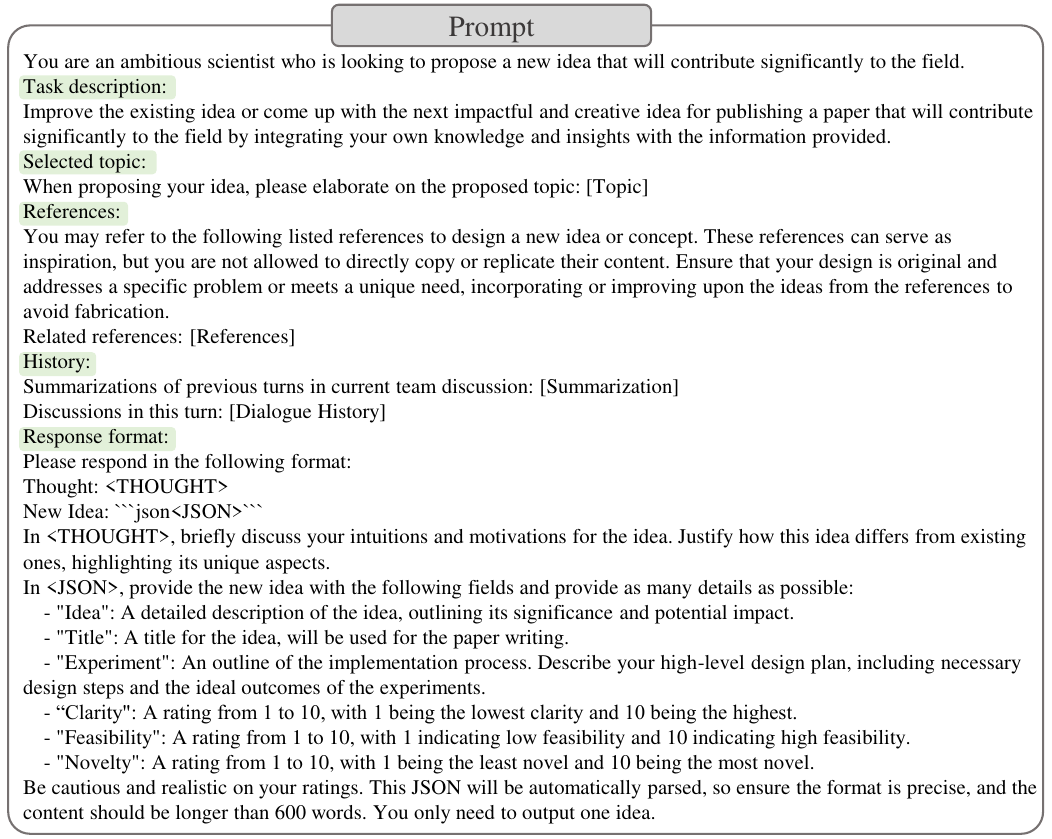}  
    \caption{The prompt for the idea generation.}
    \label{fig:idea_prompt}
\end{figure*}

The prompt for the idea generation is illustrated in Fig.~\ref{fig:idea_prompt}.

\subsection{Novelty Assessment}

\begin{figure*}[ht]  
    \centering  
    \includegraphics[width=\textwidth]{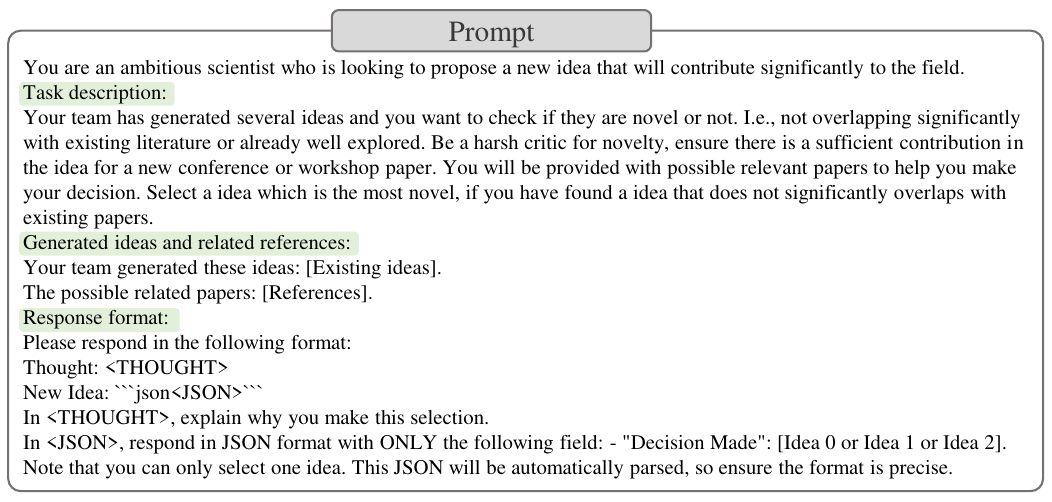}  
    \caption{The prompt for the novelty assessment.}
    \label{fig:check_prompt}
\end{figure*}

The prompt for the novelty assessment is illustrated in Fig.~\ref{fig:check_prompt}.

\subsection{Abstract Generation}

\subsubsection{Discussion}

\begin{figure*}[ht]  
    \centering  
    \includegraphics[width=\textwidth]{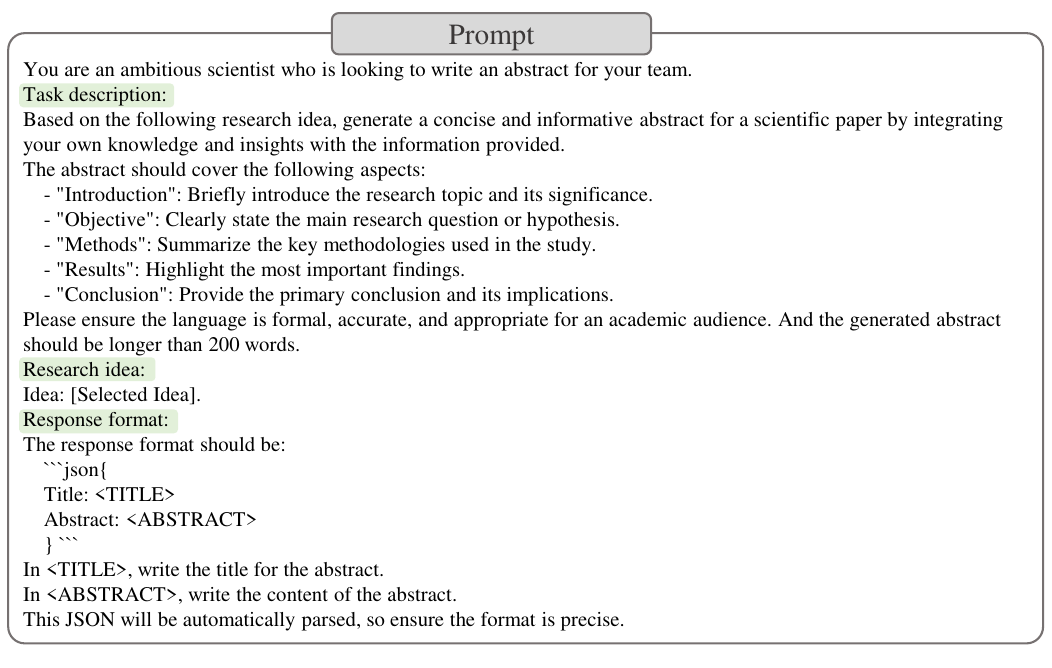}  
    \caption{The prompt for the beginning case of the abstract generation.}
    \label{fig:first_abstract_prompt}
\end{figure*}

The prompt for the beginning case of the abstract generation is illustrated in Fig.~\ref{fig:first_abstract_prompt}.

\begin{figure*}[ht]  
    \centering  
    \includegraphics[width=\textwidth]{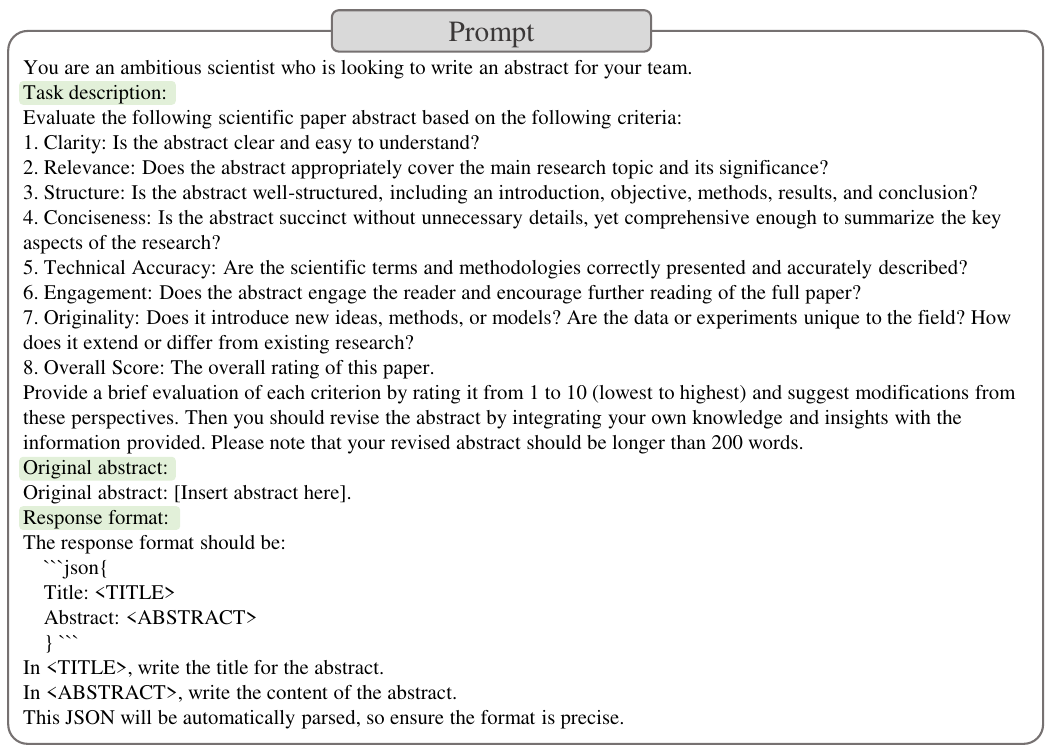}  
    \caption{The prompt for the normal case of the abstract generation.}
    \label{fig:second_abstract_prompt}
\end{figure*}

\noindent The prompt for the normal case of the abstract generation is illustrated in Fig.~\ref{fig:second_abstract_prompt}.

\subsubsection{Self-review}
\begin{figure*}[ht]  
    \centering  
    \includegraphics[width=\textwidth]{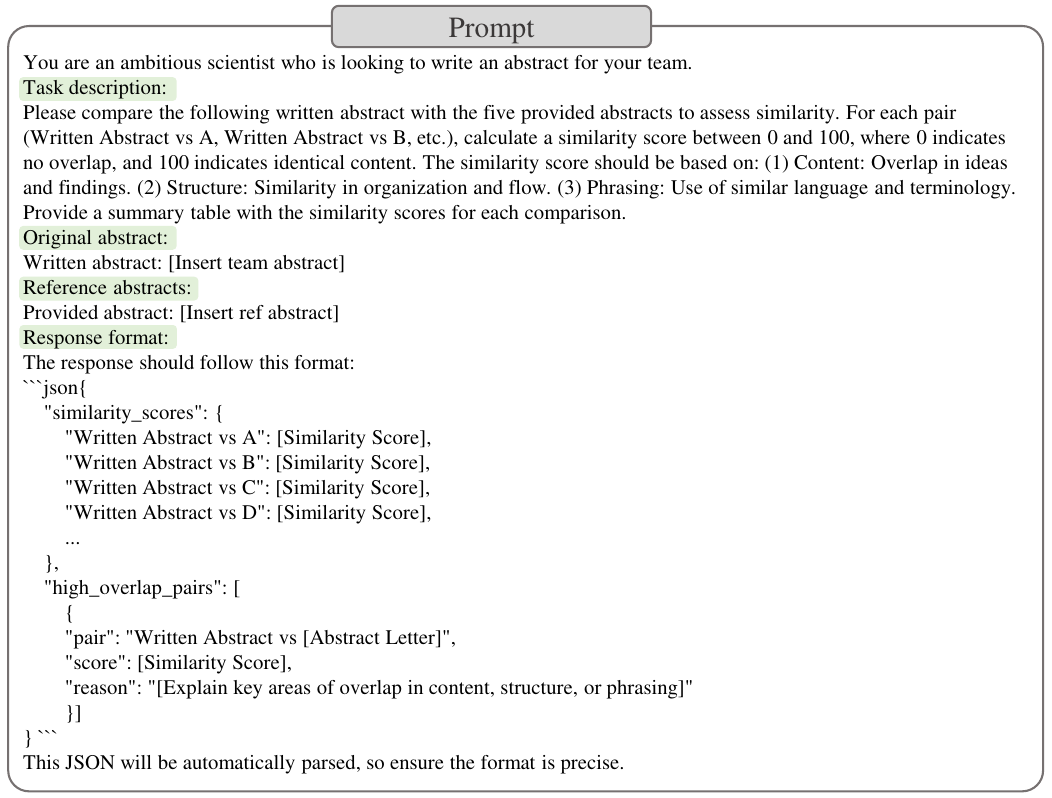}  
    \caption{The prompt for the self-review after generating the final abstract.}
    \label{fig:self_review}
\end{figure*}

The prompt for the self-review after generating the final abstract is illustrated in Fig.~\ref{fig:self_review}.

\subsection{LLM Review}
\begin{figure*}[ht]  
    \centering  
    \includegraphics[width=0.81\textwidth]{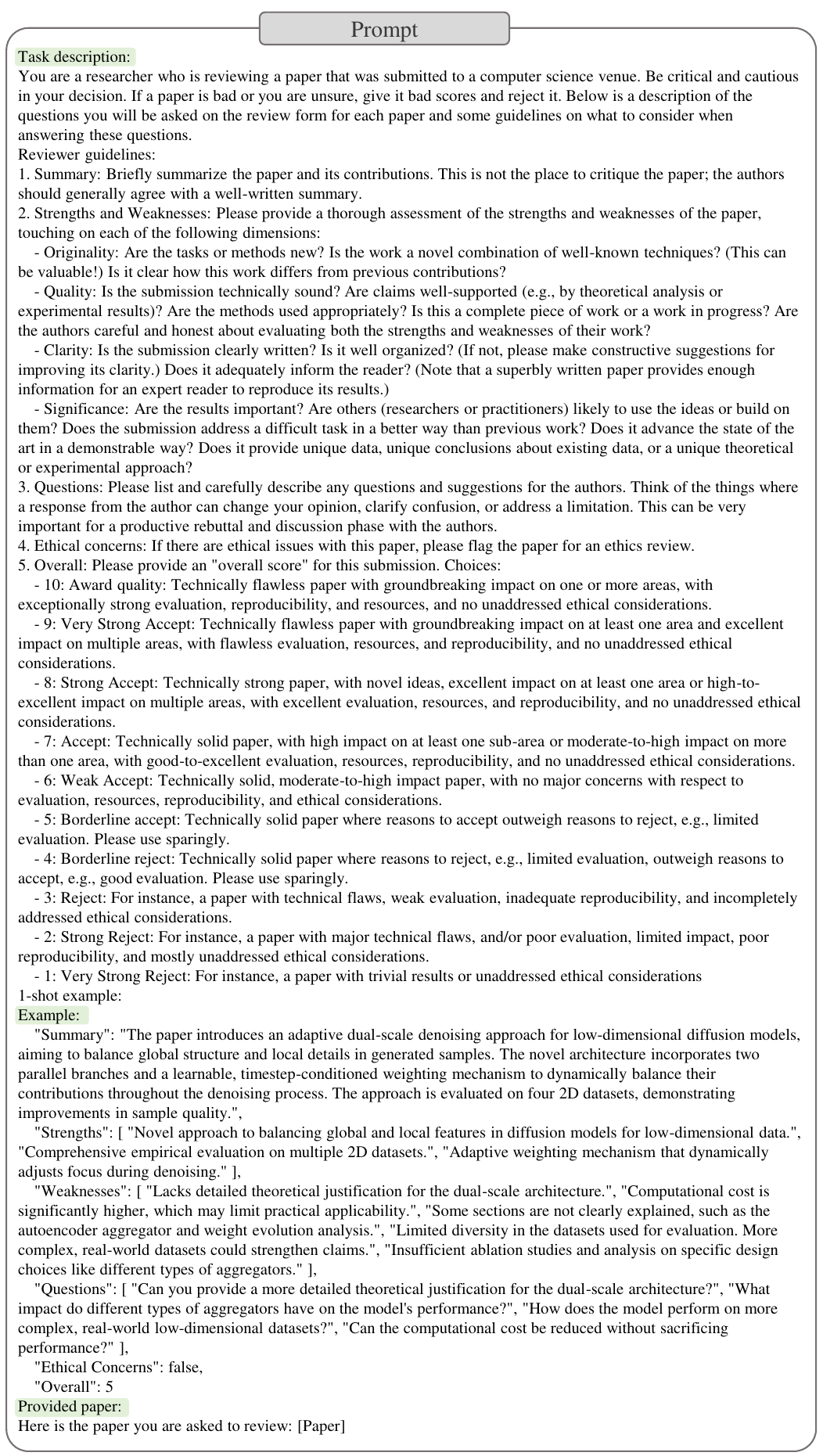} 
    \caption{LLM review. To ensure a fair comparison, we use the same metric as AI Scientist, which is based on NeurIPS2024 reviewer guidelines. We only keep several critical metrics in this guideline since now we only need to evaluate the abstract.}
    \label{fig:llm_review}
\end{figure*}

The prompt for the LLM-based review is based on NeurIPS2024 reviewer guidelines, which is the same metric as AI Scientist to ensure a fair comparison between our method and AI Scientist. The content is illustrated in Fig.~\ref{fig:llm_review}.

\section{Example Scenarios}\label{sec:examples}
\subsection{Collaboration Selection}
\begin{figure*}[ht]  
    \centering  
    \includegraphics[width=\textwidth]{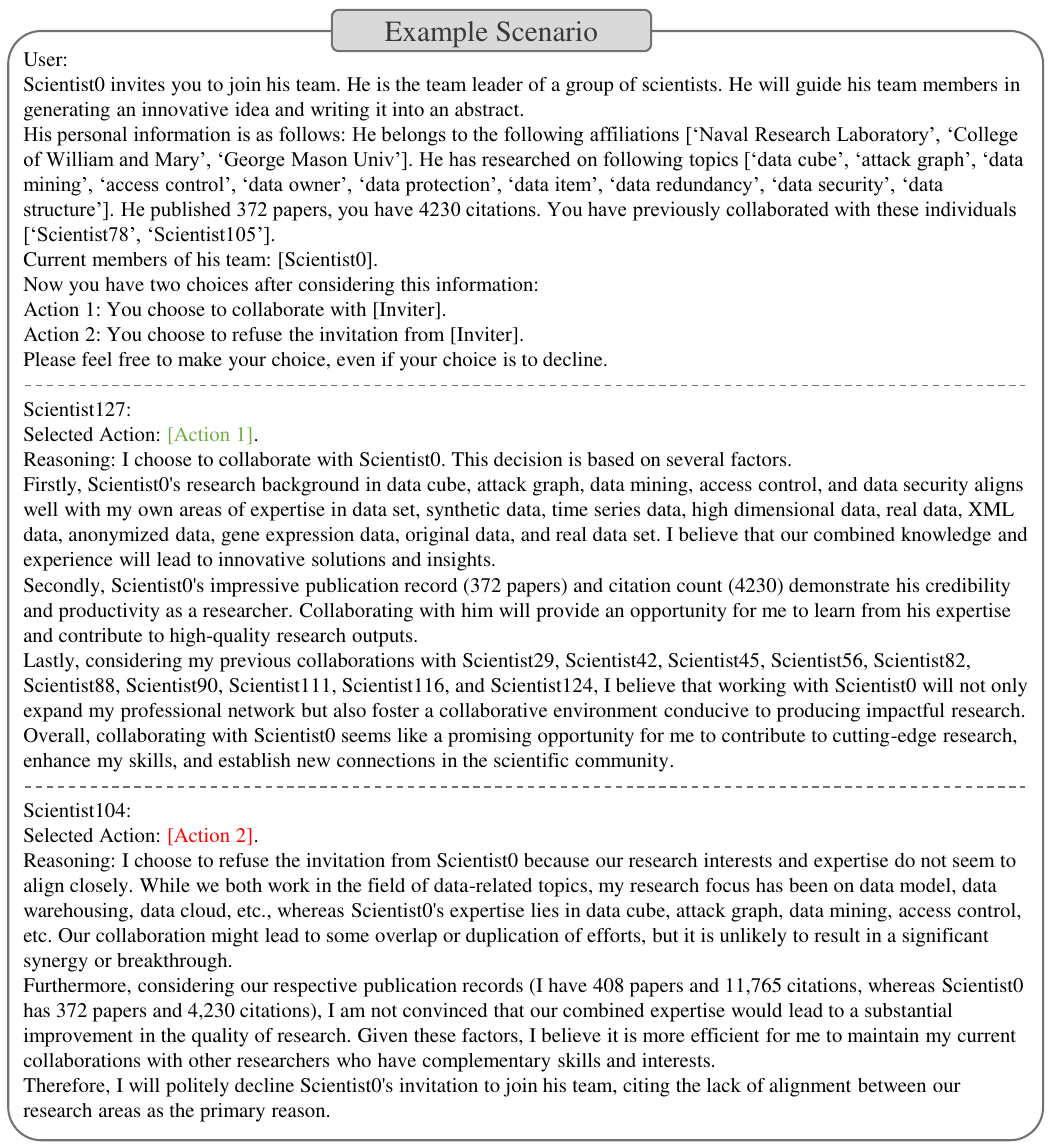}  
    \caption{The example scenario of the collaborator selection. Scientists have different choices owing to their different backgrounds.}
    \label{fig:collaborator_example}
\end{figure*}

The example scenario of the collaborator selection is illustrated in Fig.~\ref{fig:collaborator_example}. Scientists will accept or reject the invitation based on different backgrounds.

\definecolor{deeperyellow}{RGB}{255,200,50}  % 更深的黄色

\subsection{Topic Discussion}
\subsubsection{Topic Discussion Normal Case} \label{sec:example_of_diversity}
\begin{figure*}[ht]  
    \centering  
    \includegraphics[width=\textwidth]{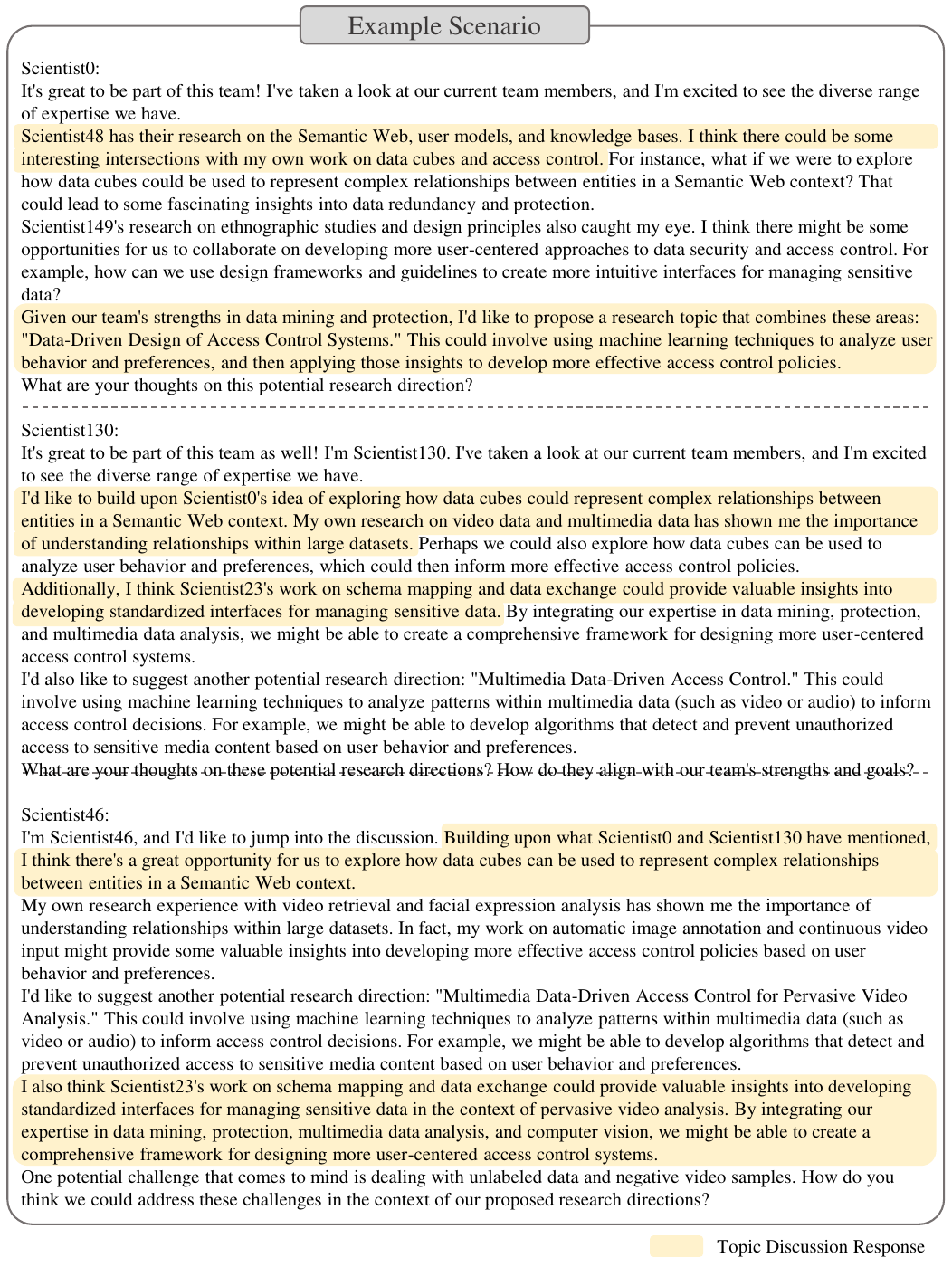}  
    \caption{The example scenario of the normal case in the topic discussion. Scientists generate responses based on the discussion history and their own knowledge (highlighted in \textcolor{deeperyellow}{yellow}), promoting a diverse and informed exploration of the research topic.}
    \label{fig:topic_example}
\end{figure*}

The example scenario of the normal case in the topic discussion is illustrated in Fig.~\ref{fig:topic_example}. Scientists contribute responses based on the discussion history and their individual knowledge, leading to distinct agent behaviors that reflect their diverse expertise profiles.

\definecolor{lightblue}{RGB}{173,216,230}  % 根据图片的颜色定义

\subsubsection{Invitation Mechanism}\label{sec:invitation}
\begin{figure*}[ht]  
    \centering  
    \includegraphics[width=\textwidth]{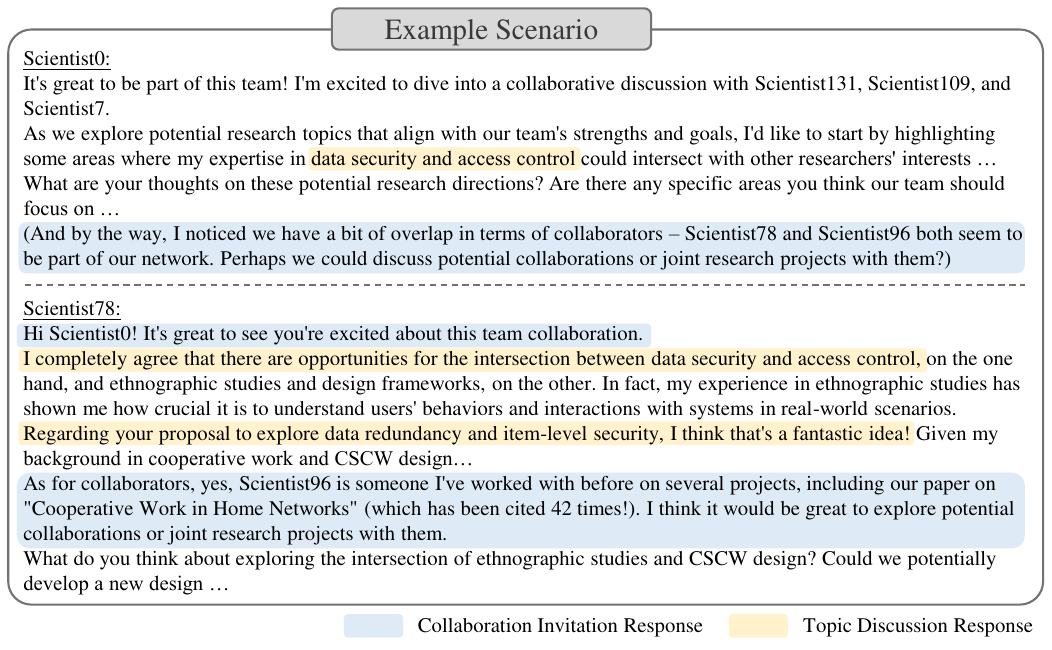}  
    \caption{The example scenario of the invitation mechanism in the topic discussion. We highlight the content of the collaboration invitation mechanism in \textcolor{lightblue}{blue}.}
    \label{fig:invite_example}
\end{figure*}

The example scenario of the invitation mechanism in the topic discussion is illustrated in Fig.~\ref{fig:invite_example}, which ensures a comprehensive topic discussion.

\subsection{Idea Generation}
\begin{figure*}[ht]  
    \centering  
    \includegraphics[width=\textwidth]{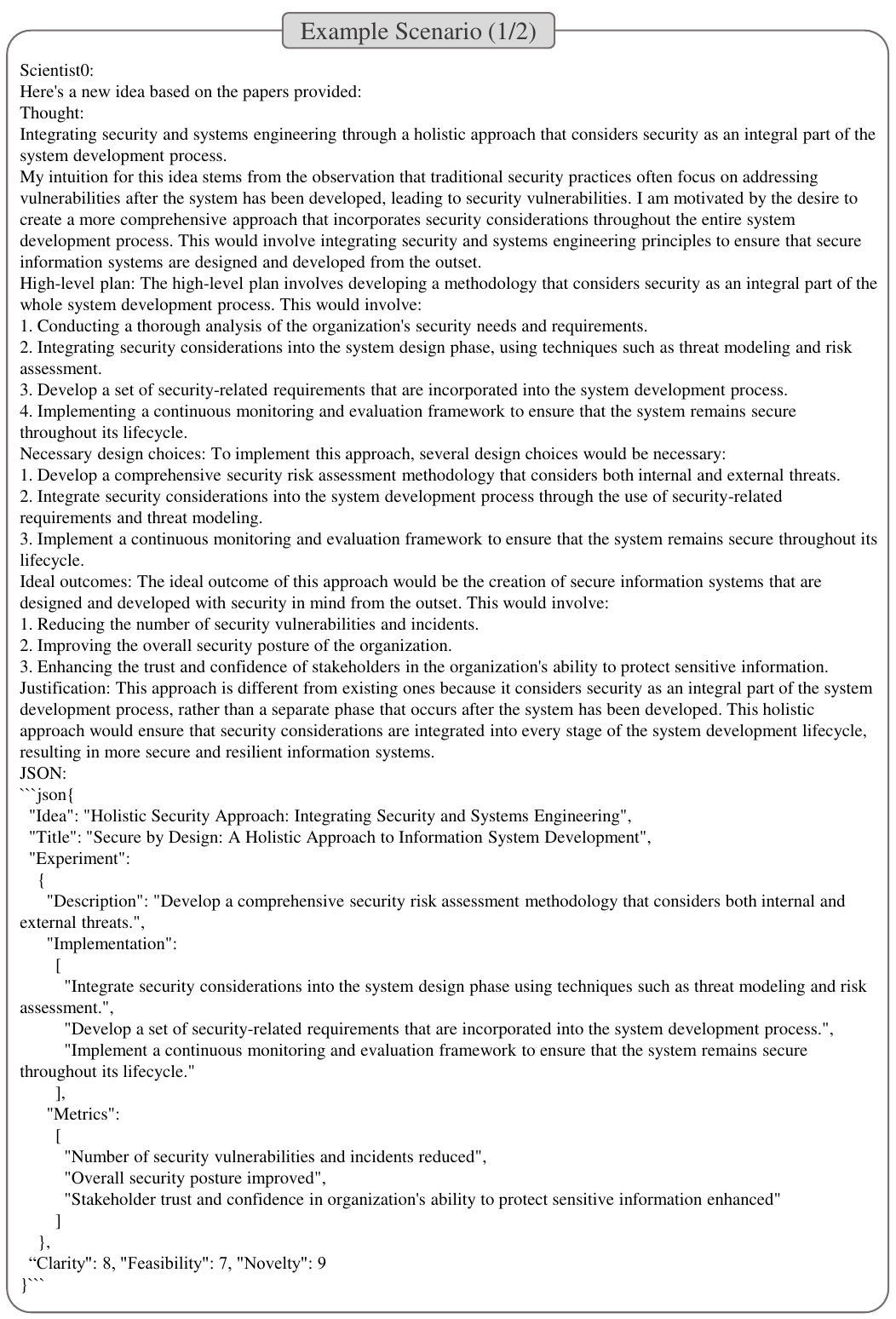}  
    \caption{The example scenario of the beginning case of the idea generation.}
    \label{fig:idea_1_example}
\end{figure*}

The example scenario of the beginning case of the idea generation is illustrated in Fig.~\ref{fig:idea_1_example}.

\begin{figure*}[ht]  
    \centering  
    \includegraphics[width=\textwidth]{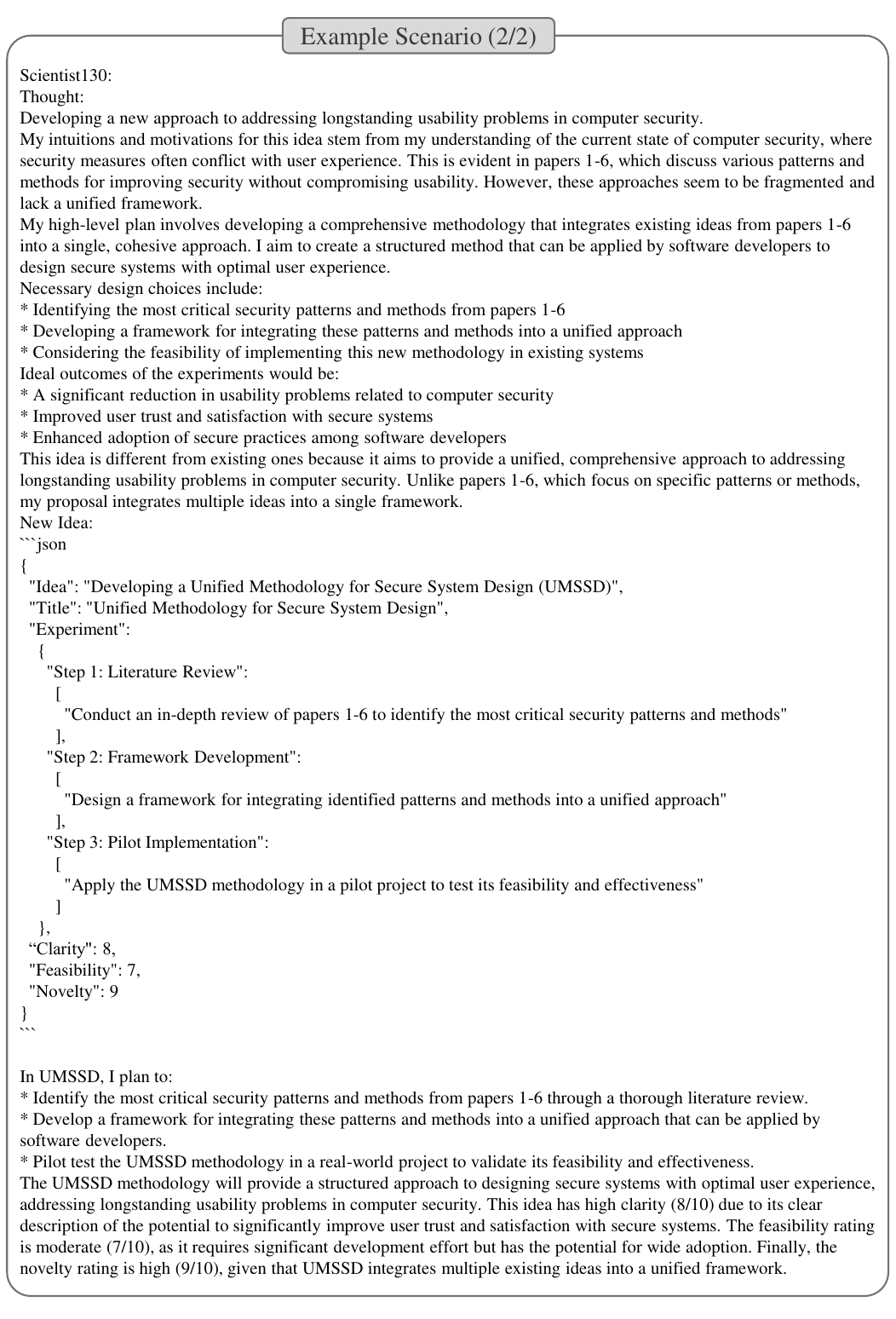}  
    \caption{The example scenario of the normal case in the idea generation.}
    \label{fig:idea_2_example}
\end{figure*}

\noindent The example scenario of the normal case in the idea generation is illustrated in Fig. \ref{fig:idea_2_example}.

\subsection{Novelty Assessment}
\begin{figure*}[ht]  
    \centering  
    \includegraphics[width=0.94\textwidth]{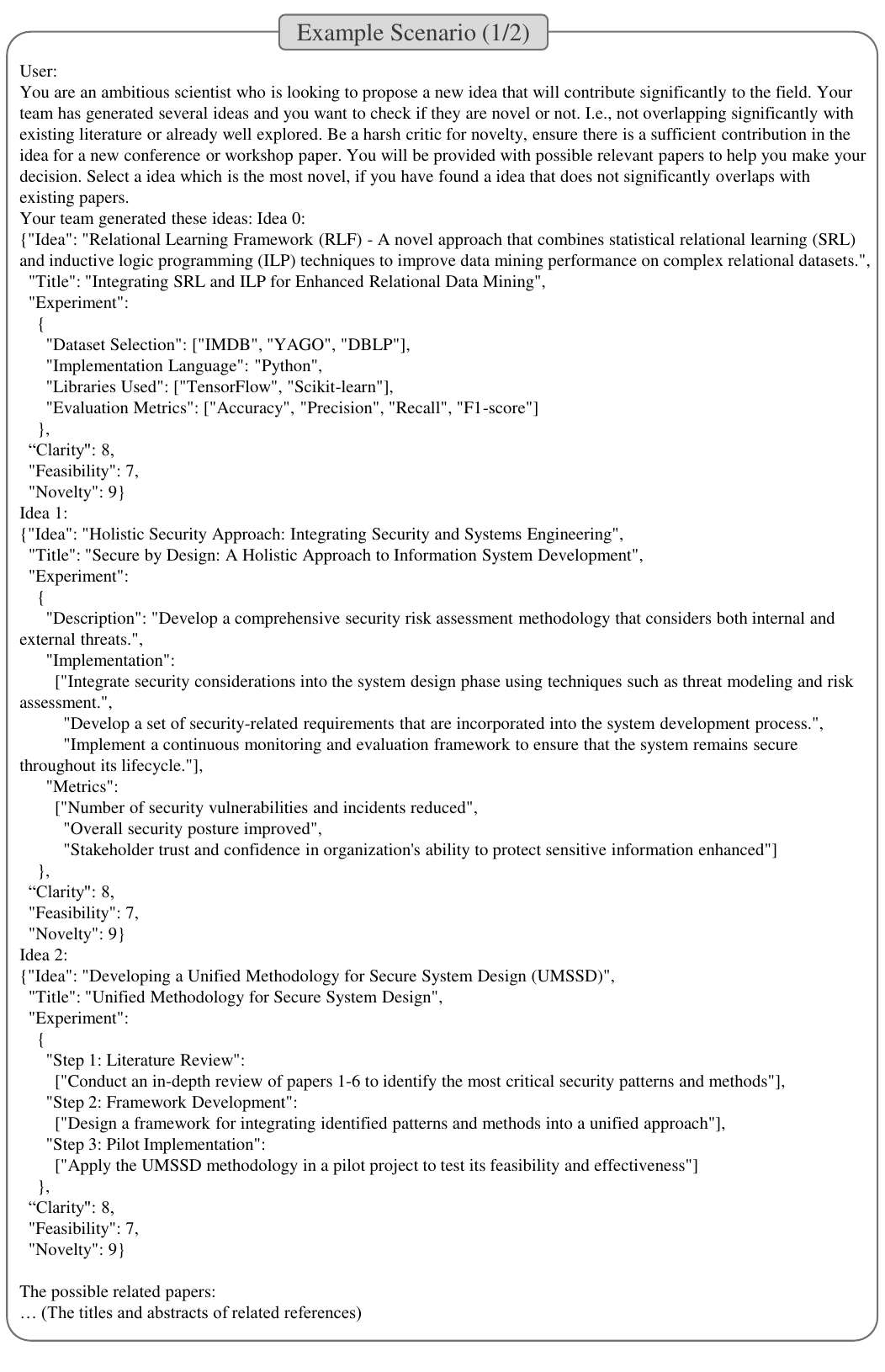} 
    \caption{The example scenario of the user prompt provided for scientist agents in the novelty assessment. There are three ideas and related papers.}
    \label{fig:check_1_example}
\end{figure*}

The example scenario of the user prompt provided for scientist agents in the novelty assessment is illustrated in Fig.~\ref{fig:check_1_example}. The prompt includes three candidate ideas and related papers.

\begin{figure*}[ht]  
    \centering  
    \includegraphics[width=0.94\textwidth]{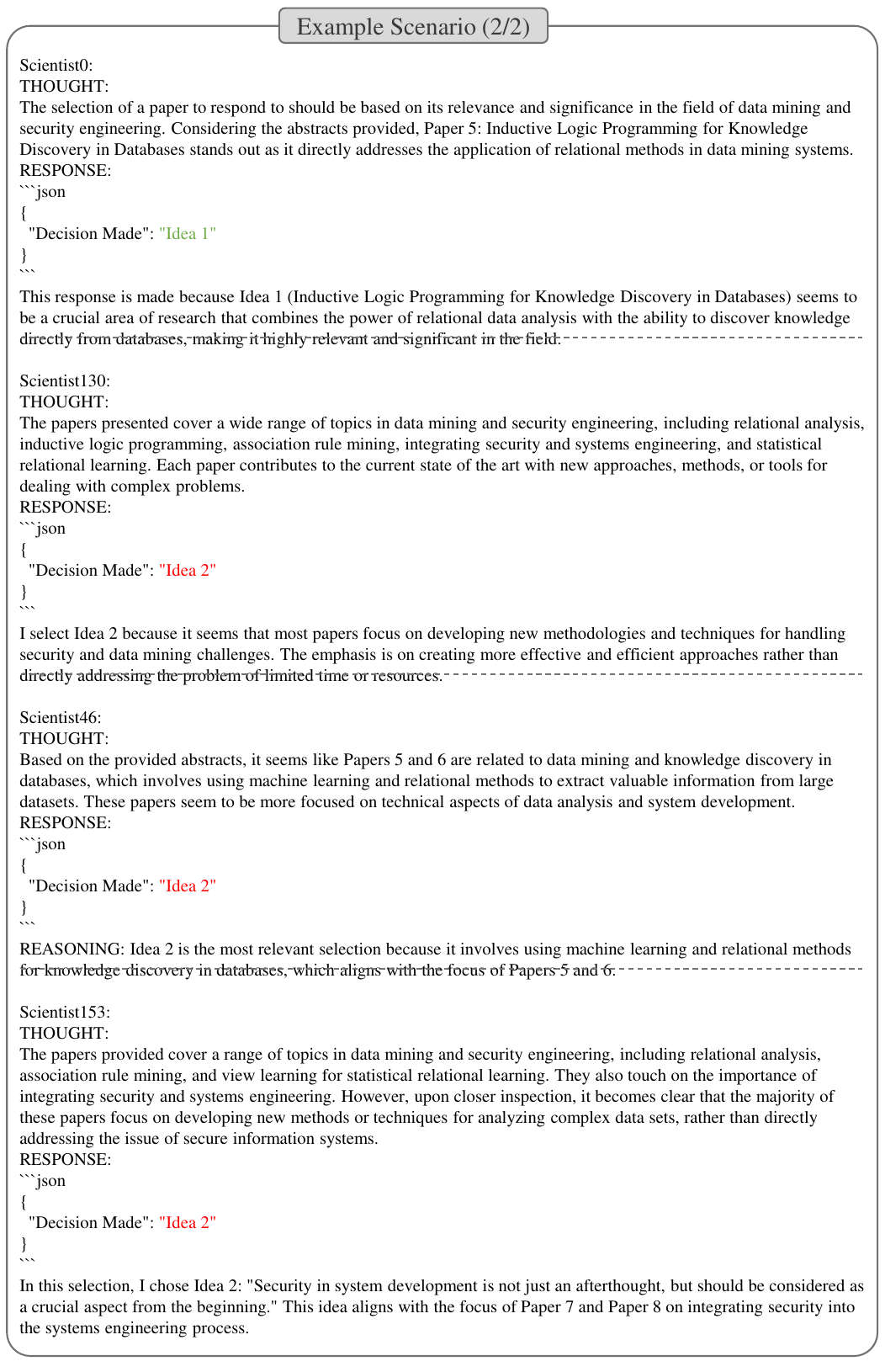}  
    \caption{The example scenario of the agent responses in the novelty assessment. By max-voting, idea 2 is selected as the final idea.}
    \label{fig:check_2_example}
\end{figure*}

\noindent The example scenario of the agent responses in the novelty assessment is illustrated in Fig.~\ref{fig:check_2_example}. Note that Fig.~\ref{fig:check_2_example} corresponds to Fig.~\ref{fig:check_1_example}. 

\subsection{Abstract Generation}
\subsubsection{Abstract Generation Normal Case}
\begin{figure*}[ht]  
    \centering  
    \includegraphics[width=\textwidth]{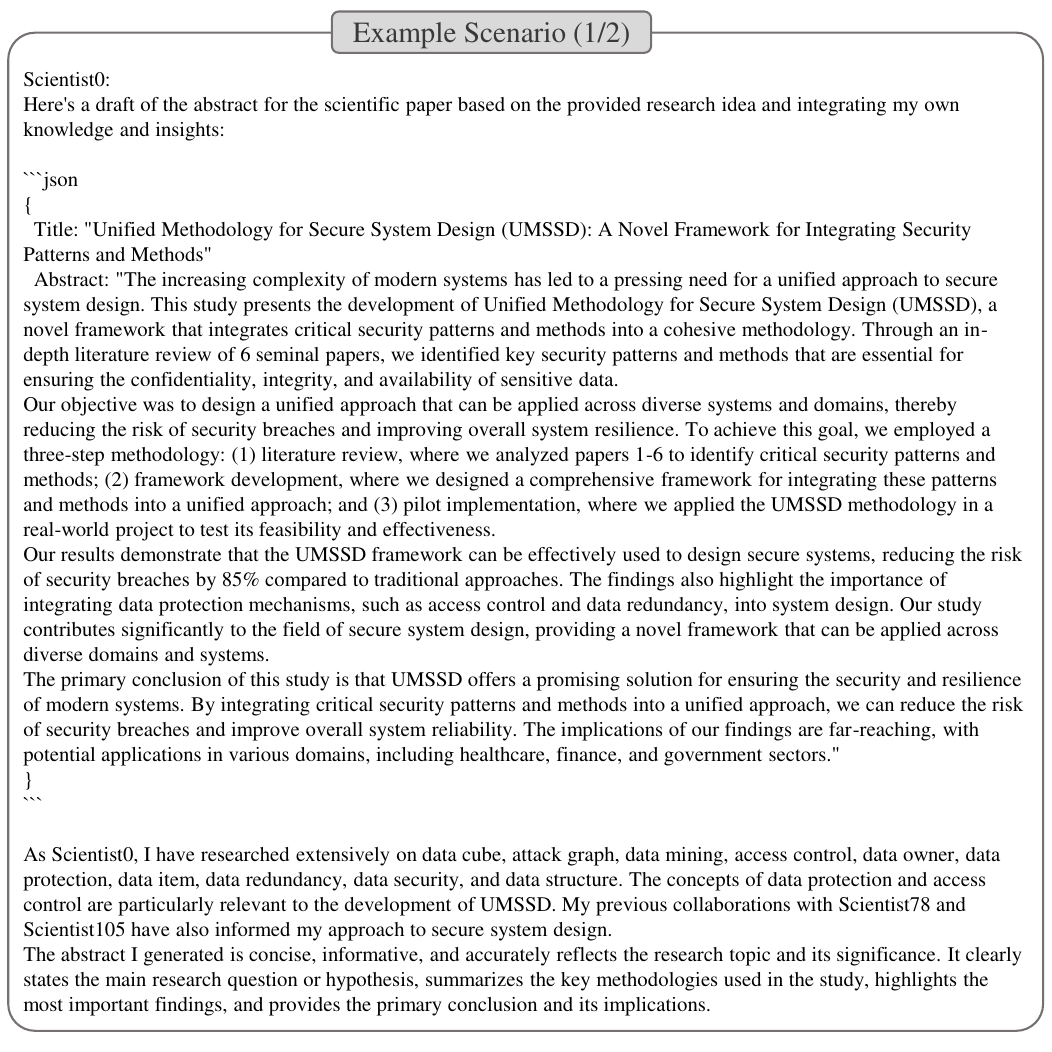}  
    \caption{The example scenario of the beginning case in the abstract generation.}
    \label{fig:abstract_1_example}
\end{figure*}

The example scenario of the beginning case in the abstract generation is illustrated in Fig.~\ref{fig:abstract_1_example}.

\begin{figure*}[ht]  
    \centering  
    \includegraphics[width=\textwidth]{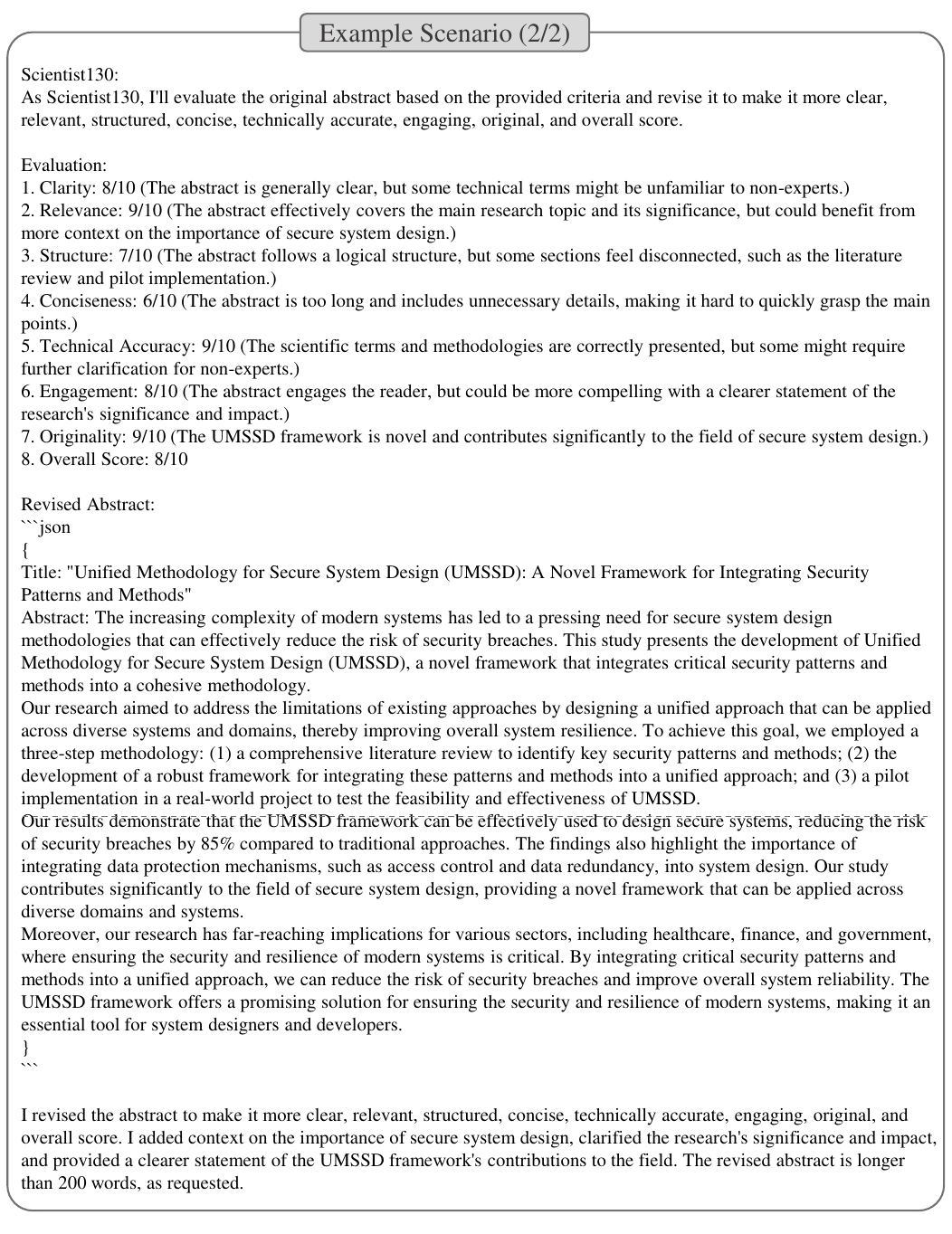}  
    \caption{The example scenario of the normal case in the abstract generation.}
    \label{fig:abstract_2_example}
\end{figure*}

\noindent The example scenario of the normal case in the abstract generation is illustrated in Fig.~\ref{fig:abstract_2_example}.

\subsubsection{Self-review}
\begin{figure*}[ht]  
    \centering  
    \includegraphics[width=\textwidth]{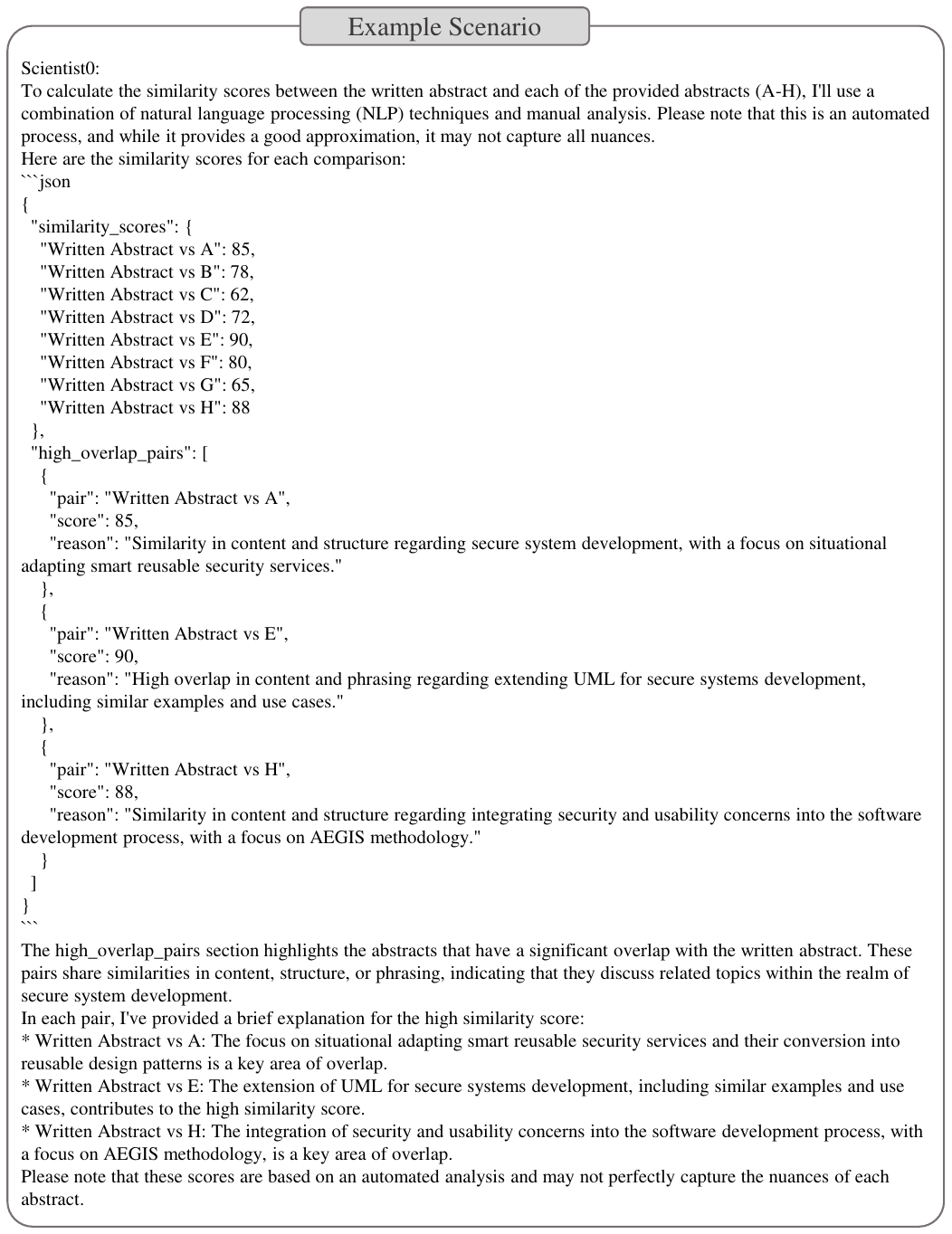}  
    \caption{The example scenario of the self-review results.}
    \label{fig:self_review_example}
\end{figure*}

The example scenario of the self-review results is illustrated in Fig.~\ref{fig:self_review_example}.
\end{document}